\documentclass[review,dvipdfmx]{elsarticle}

\usepackage{hyperref}
\journal{ }

\usepackage{amsmath}
\usepackage{amsthm}
\usepackage{amssymb}
\usepackage{eucal}
\usepackage{mathrsfs}
\usepackage{subfigure}

\usepackage{fix-cm}
\usepackage{bm}
\usepackage[dvipdfmx]{color}

\def\dO{\ \text{d}\hspace{-0.3mm} \Omega}
\def\dG{\ \text{d}\hspace{-0.3mm}\Gamma}

\def \div{\mbox{\rm div}}
\def\x{\bm{x}}

\def \RR {\mathbb R}

\def \SS {\mathbb S}
\def \PP {\mathbb P}

\def \P {\mathscr{P}}

\begin{document}

\begin{frontmatter}

\title{Geometric Shape Features Extraction\\Using a Steady State Partial Differential Equation System}

\author[mymainaddress]{T. Yamada \corref{mycorrespondingauthor}}
\ead{takayuki@me.kyoto-u.ac.jp}

\cortext[mycorrespondingauthor]{Corresponding author}
\address[mymainaddress]{Department of Mechanical Engineering and Science, Kyoto University, C3 Kyoto-Daigaku-Katsura, Nishikyo-Ku, Kyoto, 615-8504, Japan.}

\begin{abstract}
A unified method for extracting geometric shape features from binary image data using a steady-state partial differential equation (PDE) system as a boundary value problem is presented in this paper.
The PDE and functions are formulated to extract the thickness, orientation, and skeleton simultaneously. The main advantage of the proposed method is that the orientation is defined without derivatives and thickness computation is not imposed a topological constraint on the target shape.
A one-dimensional analytical solution is provided to validate the proposed method. In addition, two-dimensional numerical examples are presented to confirm the usefulness of the proposed method.
\end{abstract}

\begin{keyword}
Partial differential equations; Geometric shape features; Shape analysis; Finite element method; Computer aided engineering
\end{keyword}

\end{frontmatter}

%\linenumbers
\section{Introduction}
The development of remarkable image analysis technology in recent years has helped address several problems in various fields, such as materials science \cite{yamashita2014volume}, mechanical engineering \cite{benkHo2001algorithms}, biomechanics \cite{sera2003three,zhenjiang2000zernike}, medicine \cite{hildebrand1997new,hutton2008voxel}, and shape analysis\cite{kokaram2003digital,costa2000shape}.
For example, the skeleton can be extracted from computed tomography (CT) and magnetic resonance imaging (MRI) data, contributing to an understanding of its structure. In particular, the estimation of local thickness is an important measure for disease propagation.
In reverse engineering of mechanical products \cite{fujimori2005surface}, the extraction of geometrical features (e.g., curvature and edge information) from X-ray CT images is an important analytical technique when designing and developing novel high-performance systems in a short time.
When designing mechanical products, the extraction of members that exceed the allowed minimal thickness in computer-aided design (CAD) models is an important design consideration. 
Therefore, feature extraction is used in a variety of tasks in the fields of computer vision, image processing, and digital engineering.

This paper presents a unified method for extracting geometric features by using a partial differential equation (PDE).
In the following section, related research on feature extraction and PDE-based image processing is briefly discussed. Second, the basic concept and an overview of the proposed method are discussed by comparing related research with the proposed method.
Next, a PDE for geometric shape feature extraction is formulated. The shape feature functions for thickness, orientation, and skeleton are formulated based on the proposed PDE. That is, these geometric features are represented as a function of the solution of the PDE. In addition, a numerical algorithm for the proposed method based on the finite element method (FEM) is presented.
In Section \ref{sec:1d}, the validity of the proposed method is discussed based on a one-dimensional analytical solution. Finally, to confirm the validity and utility of the proposed method, several numerical examples are provided for two-dimensional cases.
%--------------------------------
%--------------------------------
%--------------------------------
\section{Related Works} \label{sec:RelatedWorks}
The tensor scale is a measure of shape features that represents thickness, orientation, and anisotropy \cite{saha2005tensor, andalo2010shape}. 
The measure defines the parameters of the largest ellipse within the target domain at each pixel point. Although the measure in the proposed method represents several geometric features simultaneously, it incurs high computational cost because the Euclidean distance is computed for each point.

Carne et al. \cite{crane2013geodesics} proposed a PDE-based method for distance computation. The basic concept of their method involves considering a fictitious heat diffusion equation with Dirichlet boundary conditions for a short time.
Then, the distance is approximated using the fundamental solution of the fictitious temperature. Although the linear diffusion equation provides a smooth solution and low computational cost, the heat effect reaches infinite distance even in a short time. Therefore, the method cannot be used to derive the exact distance from a theoretical perspective.

Related research on feature extraction is briefly discussed as follows.

%---------------------
%   Thickness
%---------------------
{\bf Thickness: } 
One significant application of thickness extraction is the determination of bone thickness \cite{saha2004measurement,liu2014robust}. 
The local thickness defined in \cite{hildebrand1997new} is estimated based on a fuzzy distance transformation \cite{saha2002fuzzy} in these methods. 
The process requires sampling depth at axial voxels that are computed by skeletonization techniques\cite{arcelli2011distance,saha1997new}.
Another challenging application of thickness measurement is the estimation of the cortical thickness of the human brain from MRI data \cite{hutton2008voxel,clarkson2011comparison}.
These methods are classified as surface-based, voxel-based, and hybrid methods.
In these methods, an image is separated into three domains: grey matter, white matter, and cerebrospinal fluid.
%--- SURFACE
As discussed in \cite{clarkson2011comparison}, the surface-based method \cite{davatzikos1996using} uses a generated mesh on one side surface. Next, the mesh is deformed to fit the pair surface under a topological constraint. In general, calculation of the advection requires a high computational cost to ensure consistent topology \cite{han2004cruise}.

%--- VOXEL
In contrast, voxel-based methods can be categorized into morphological, line integral, diffeomorphic registration, and Laplacian-based methods.
%  Morphological
The morphological method \cite{lohmann2003morphology} divides each voxel into inner and outer domains. The thickness is computed using a Euclidean distance transformation.
%- Line Integral
In line integral-based approaches  \cite{aganj2009measurement,scott2009fast}, every line integral centered at each point is computed and the minimum value is defined as the thickness.　
%-- diffeomorphic registration
Diffeomorphic registration \cite{das2009registration} also requires calculating surface deformation. 
%--  Laplasian based method
Jones et al. \cite{jones2000three} proposed the Laplacian-based approach. In this approach, two surfaces consisting of a target shape are considered. It is assumed that the surfaces are topologically equivalent. The Laplace equation is considered in the domain surrounded by the two surfaces, where Dirichlet boundary conditions with different constant values are imposed. The thickness between these surfaces is defined as the length along the normal direction of the isosurface of the potential field.
Based on this approach, Yezzi et al. \cite{yezzi2003eulerian,yezzi2001pde} proposed the Eulerian approach to directly compute the thickness along the normal direction. Hybrid Eulerian--Lagrangian approaches have also been proposed \cite{rocha2005hybrid,acosta2009automated}. The multiple Laplace equation has been used for time-dependent estimation problems \cite{cardoso2011extraction}.
The main advantage of PDE approaches is that thickness is uniquely defined at any point.
However, this basic idea is restricted under the topological constraint. Furthermore, the inner subsurface must be distinguished from the outer subsurface.

% Proposed method
Although, the proposed method is similar to the Laplacian-based approach,　it can be applied to voxel-based and surface-based data because the proposed PDE can be easily solved using the boundary element method.
In addition, the proposed method essentially overcomes the topological constraints and does not require dividing surfaces into an inner and outer surface.
Furthermore, the proposed PDE is well posed, that is, the solution is unique and numerically stable.

%---------------------
%   Skeleton
%---------------------
{\bf Skeleton: } 
The skeleton function \cite{montanari1968method,blum1978shape} can be applied in a wide range of fields, such as medical science, animation, and reverse engineering.
As discussed in \cite{cornea2007curve}, these methods are categorized into topological thinning methods \cite{saha1997new,palagyi19983d}, methods using a distance field \cite{arcelli2011distance,bitter2001penalized}, geometric methods \cite{amenta2001power}, and methods using a generalized potential field model \cite{abdel1994multiresolution}.
The proposed method is closest to the generalized potential field model \cite{ahuja1997shape,grigorishin1998skeletonisation}.
This model requires considering a fictitious electrostatic potential field with sources on the surface. 
The main advantage of this method is that it can provide relatively good results. However, high computational costs are incurred because the Newton potential field is computed superpositioning each point. In addition, these algorithms do not consider numerical stability from a mathematical perspective.
Several approaches have been proposed to overcome problems related to connectivity and robustness in recent years.
For instance, the erosion thickness approach provides a robust, connected skeleton \cite{yan2016erosion}. 

Aslan et al. \cite{aslan2008disconnected} proposed a disconnected skeleton based on the distance function with the heat diffusion equation under a nonzero Dirichlet boundary condition. Aubert et al. \cite{aubert2014poisson} used a heat diffusion equation with a constant heat source to extract the distance function and skeleton.
Gao et al. \cite{gao20182d} presented connected skeleton extraction based on the heat diffusion equation. These heat equation-based methods require precise extraction of ridge curves and calculation of heat diffusion from boundaries of short times because the Dirichlet boundary condition is imposed at the shape boundaries.

In addition, PDE-based approaches are used in many related fields; these are described in the following section.
%---------------------
%   Image processing
%--------------------

{\bf Image Processing: }
An elliptic PDE is extensively used in image processing. 
Poisson image editing \cite{perez2003poisson} is an image-editing method that requires solving the Poisson equation. The process requires preserving the gradient of a source image for seamless image editing. Pixels with a high gradient are extracted by using the Poisson equation.
Poisson matching \cite{sun2004poisson} also uses the Poisson equation for image matching.
This basic concept of these image processing methods is related to the extraction of geometric shape features. 
%---------------------
%   topology optimization
%--------------------

{\bf Topology optimization: }
The use of PDE-based feature evaluation has been proposed for topology optimization \cite{sato2017manufacturability}.
The manufacturability in molding is evaluated by superposition of the solution to the PDE.
The main advantages of this optimization procedure are the shape and topological sensitivities that are derived using the adjoint variable method without limiting the design space.
%---------------------
%   
%---------------------
\section{Concept and Overview}\label{sec:concept}
\begin{figure*}[htb]
	\begin{center}
		\includegraphics[width=12cm]{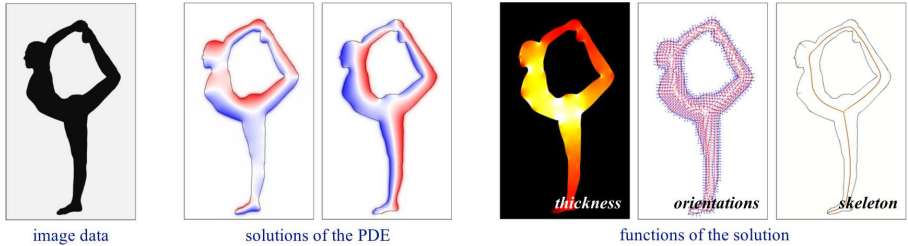}
		\caption{Overview of the proposed method: the proposed method extracts geometric shape features, such as thickness, orientation, and skeleton using the solutions of a proposed linear partial differential equation, whose coefficients are given by image data. The image on the left shows input image data that are used to determine the coefficients. The images in the middle are solutions to the linear partial differential equation. The three images on the right show various solutions.}
		\label{fig:concept}
	\end{center}
\end{figure*}
The basic concept of shape feature extraction via the use of a steady-state PDE is involves the extraction of target geometric features, such as thickness, skeleton, orientation, and curvature, from the target image as a function of the solution to the PDE system, as shown in Figure \ref{fig:concept}.
This paper presents a formulation of a steady-state PDE system and functions for basic geometric features represented by a solution to the PDE system. 

The proposed method has the following advantages:
\begin{enumerate}
	\item Multiple geometric features are computed simultaneously by solving the steady-state PDE system. 
	\item Relatively small shape fractionation on the surface is automatically neglected in the diffusion effect, that is, the method automatically inherits the numerical advantages of the method used to solve the PDE.
	\item Thickness extraction does not require any topological constraints and distinction between inner and outer surfaces. 
	\item The formulation of the PDE and geometric shape feature functions are not dependent on dimensions.  Conditions on the shape boundaries are not imposed from a numerical perspective.
\end{enumerate}

%---------------------
%  
%---------------------
\section{Formulation}\label{sec:formulation}
\subsection{Partial differential equation for geometric shape features extraction}
\begin{figure}[ht]
	\begin{center}
		\includegraphics[width=5cm]{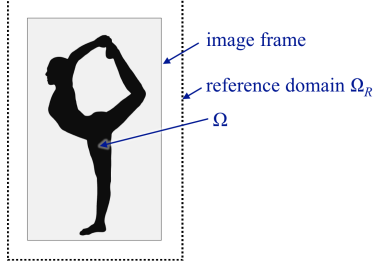}
		\caption{Definitions for formulation}
		\label{fig:def}
	\end{center}
\end{figure}

First, a steady-state linear PDE system is defined for extracting geometric shape features of a binary image.
A reference domain $\Omega_R$ is considered that consists of a black domain $\Omega$ and a white domain $\Omega_R \setminus \Omega$ whose digital signals are $1$ in the black domain and $0$ in the white domain, respectively. It is assumed that the reference domain $\Omega_R$ contains the target image, as shown in Figure \ref{fig:def}.
Here, extraction of shape features from the black domain $\Omega$ is considered. Note that the shape features in the white domain can be considered in a situation with the opposite signal.
This study focuses on the similarity of shape features when optimizing periodic homogenization \cite{allaire2018optimization}. The PDE system is formulated as follows:
\begin{equation}
\begin{cases}
- \div \Bigl(  \tilde{a} \nabla s_i -e_i \chi \Bigl) +\alpha(1-\chi) s_i=0\qquad &\text{in}\quad \Omega_R \\
s_i  =0\qquad &\text{on}\quad \partial \Omega_R
\end{cases}
\label{eq:PDE}
\end{equation}
where $s_i\in H^1(\Omega_R)$ is the $i$-th state variable, $e_i$ is the canonical base of $\RR^d$, $\tilde{a} \in \RR_+$ is the diffusion coefficient, $\alpha  \in \RR_+ $ is the damping coefficient, and $d$ is the dimension. The coefficient $\alpha$ is set to a relatively large value in order to decrease the mutual effect via the white domain.  The characteristic function $\chi \in L^{\infty}(\Omega_R)$ is defined as
\begin{equation}
\chi (\x) :=
\begin{cases}
&1\qquad {\rm for} \quad \x \in \Omega\\
&0\qquad {\rm for} \quad \x \in  \Omega_R \setminus \Omega.
\end{cases}	
\end{equation}
The characteristic function is equivalent to using binary data in the target image.
Note that the inner and outer surfaces do not have to be differentiated because domains are distinguished by the characteristic function in the same manner as the topology optimization method \cite{yamada2010topology}.
In addition, the proposed PDE does not require any topological restriction. 

Next, a parameter $a$ is introduced in order for the diffusion coefficient to be satisfied as $\tilde{a}:= a h_{0}^{2}$, where $h_0>0$ is the characteristic length of the target shape size. The concept of the characteristic length is the same as that used in mechanics, that is, the non-dimensional equation and feature functions defined below are generally reasonable.
The damping coefficient $\alpha$ is defined as follows:
\begin{equation}
\alpha :=\frac{4}{a}
\end{equation}
Then, the proposed PDE contains the non-dimensional diffusion parameter $a$. The parameter $a$ should be set to be sufficiently small because the damping coefficient $\alpha$ is defined to prevent effects from the surrounding domain and the boundary of the reference domain $\partial \Omega$, that is, the damping coefficient must be set to a large value in order to make the state variables $s_i$ to zero nearly everywhere in the white domain. 
The number of potential fields defines the dimension, that is, feature extraction is required for a vector field consisting of independent potential fields. 

To physically interpret the proposed PDE, the weak and strong forms are derived by introducing a vector field $\bm{s}=[s_1, s_2,...,s_n]^T$. The weak form is derived as follows:
\begin{equation}
\begin{cases}
- \div \Bigl(  \tilde{a} \nabla \bm{s} -\rm{\bm{Id}} \Bigl) =0\qquad &\text{in}\quad \Omega \\[10pt]
- \div \Bigl(  \tilde{a} \nabla \bm{s}  \Bigl) +\alpha \bm{s}=0\qquad &\text{in}\quad \Omega_R \setminus \Omega \\[8pt]
\bm{s}  =0\qquad &\text{on}\quad \partial \Omega_R
\end{cases}
\label{eq:weak}
\end{equation}
As shown in Equation (\ref{eq:weak}), the black domain is governed by the diffusion equation, as with the steady state linear elastic equation. The source is shown in the black domain and its magnitude is the divergence of the identity matrix.
The damping term decreases the mutual effect via the white domain because the Helmholtz equation governs the behavior of the system in this domain.
Therefore, the state variable vector $\bm{s}$ exponentially converges to the zero vector in the white domain.
The strong form is derived as follows:
\begin{equation}
\int_{\Omega_R} \Bigl(  \tilde{a} \nabla \bm{s} : \nabla \bm{\xi} \Bigl)\dO 
+\alpha\int_{\Omega_R \setminus \Omega}  \bm{s} \cdot \bm{\xi} \dO 
=\int_{\partial \Omega} \bm{n} \cdot \bm{\xi} \dG
\end{equation}
where $\bm{\xi} \in H^{1}_{0}(\Omega_R)^d$ is a test function. The left and right sides are bilinear terms and the source term, respectively. Fictitious traction is applied along the normal direction on the surface $\partial \Omega$ with unit magnitude. Therefore, the state variable vector $\bm{s}$ lies along the normal direction of the shape
because the damping term is relatively large in the left-hand side without domain around the surface $\partial \Omega$.
Thus, fictitious traction is not directly imposed on the surface $\partial \Omega$.

\subsection{Shape feature tensor $\SS^*$}
The shape feature tensor $\SS^*$ is defined as follows:
\begin{equation}
\SS_{ij}\left(\{s\}_{1\le i \le d}\right) := 
\frac{1}{2}
\left(
\frac{\partial s_i}{\partial x_j}
+
\frac{\partial s_j}{\partial x_i}
\right)
\end{equation}
The key geometric features are defined using the shape feature tensor $\SS$  because the tensor includes all directions of the gradient with respect to the vector $\bm{s}$. Note that the definition of the gradient tensor is not unique. For instance, a strain tensor in linear elastic dynamics includes transpose components.
The eigenvalues $\lambda_{s}^{(i)}$ of the shape feature tensor matrix and normalized eigenvector $\x_{s}^{(i)}$, where the order of eigenvalues are defined to satisfy $\lambda_{s}^{(i)}\le \lambda_{s}^{(i+1)}$ are introduced.
In addition, the state variable $\tilde{s}_i$ using the eigenvector is defined as follows:
\begin{align}
\left(
\begin{matrix}
\tilde{s}_1\\
\tilde{s}_2\\
\vdots\\
\tilde{s}_d
\end{matrix}
\right)
:= 
\left( \bm{x}_{s}^{1}\quad \bm{x}_{s}^{2}\quad \cdots\quad  \bm{x}_{s}^{d}\right)^T
\left(
\begin{matrix}
s_1\\
s_2\\
\vdots\\
s_d
\end{matrix}
\right)
\end{align}
\subsection{Thickness function}
The thickness is inversely proportional to the sum of the derivatives of the state variables along each direction of the canonical base. That is, the following inverse thickness function $f_h$ is inversely proportional to the local thickness of the target shape:
\begin{align}
f_h\left(\{s\}_{1\le i \le d}\right)
:&=h_{0}^{2}\left(\sum_{i=1}^{d}  \frac{\partial s_i}{\partial x_i} \right) \chi\\
&= h_{0}^{2}  \left(\sum_{i=1}^{d}  \lambda_{s}^{(i)} \right)\chi
\end{align}  
The detailed properties of the inverse thickness function $f_h$ are discussed in Section \ref{sec:2d}.
Using the property with respect to thickness, the thickness function $h_f$ is defined as follows:
\begin{align}
h_f\left(\{s\}_{1\le i \le d}\right):&=
h_0\left\{ 
\frac{1}{f_h\left(\{s\}_{1\le i \le d}\right) }
- a
\right\}
\chi.
\end{align}
The value of the thickness function $h_f$ represents the local thickness in the black domain $\Omega$.
\subsection{Orientation vector function}
As discussed in the weak formulation of the PDE, the state variable vector represents the direction normal to the shape. Therefore, the orientation with respect to the normal direction is expressed as follows:
\begin{equation}
{\bm n_f} \left(\{s\}_{1\le i \le d}\right):=
\frac{1}{\displaystyle \sqrt{\sum_{i=1}^{d} s_{i}^{2}}}
\left(
\begin{matrix}
&s_{1}\\
&s_{2}\\
&\vdots\\
&s_{d}
\end{matrix}
\right)
\end{equation}
The tangential orientation vector $\bm{t}_f$ is computed by applying a rotational transformation to the normal orientation vector $\bm{n}_f$.
\subsection{Skeleton function}
One of the skeleton functions is formulated as follows:
\begin{align}
f_s \left(\{s\}_{1\le i \le d}\right)&:=
\P \left( 
\frac{
	\sqrt{\tilde{s_1}^2} }{\lambda_1}
\right)
\chi.
\end{align}
where $\P$ is a pulse function defined as
\begin{align}
\P (x)=
\begin{cases}
&0\qquad {\rm if} \quad -w >x\\
&1\qquad {\rm if} \quad -w\le x \le  w\\ 
&0\qquad {\rm if} \quad w < x\\
\end{cases}
\end{align}
The parameter $w>0$ for width in the nonzero domain should be defined to obtain the expected width, e.g., the pixel size. 
The function estimates the medial surfaces in three-dimensions. Therefore, another function may be defined based on the requirements of each application.
Note that the function describes a disconnected skeleton \cite{aslan2008disconnected}, and this heuristic formulation requires precise discussions.
%------------------------------------------------------------------
%--------------------------------------------------------------------------------------
% Section 
%--------------------------------------------------------------------------------------
%--------------------------------------------------------------------------------------
\section{Numerical Implementation}\label{sec:imple}
The computational procedure, which is essentially the same as the finite element method, is as follows:
\begin{enumerate}
	\item The reference domain is defined within the target shape. In general, the reference domain surrounds the input image data.
	\item The reference domain is composed of discretized finite elements; their material properties are defined based on the characteristic function $\chi$, which is defined in the input image data.
	\item The PDE system (\ref{eq:PDE}) is solved using the finite element method. That is, the numerical solution of the state variables $s_i (\x)$ are given. 
	\item The target geometric shape feature is computed from the state variables $s_i (\x)$. 
\end{enumerate}
This procedure can be easily implemented using a finite element analysis software. The numerical examples shown in Sections \ref{sec:2d} and \ref{sec:3d} are solved using COMSOL Multiphysics. 
The boundary element method is also useful for analyzing the proposed PDE if the input data format is surface data, such as STL data.
%--------------------------------------------------------------------------------------
% Section 
%--------------------------------------------------------------------------------------
%--------------------------------------------------------------------------------------
\section{Analytical validation in one dimension}\label{sec:1d}
\begin{figure}[ht]
	\begin{center}
		\includegraphics[width=7cm]{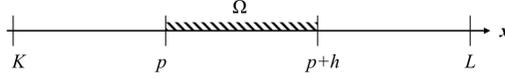}
		\caption{An isolated domain in one-dimension}
		\label{fig:NS1-FDD}
	\end{center}
\end{figure}

The analytical solutions of the proposed PDE are easily derived in one dimension.
A one-dimensional case is considered to verify the proposed method.
The distribution of the black domain shown in Figure \ref{fig:NS1-FDD} is considered.
The black domain exists between $x=p$ and $x=p+ h $. 
The boundary condition is imposed at $x=K$ and $x=L$. These positions are sufficiently far from the black domain $\Omega$.
The governing equation and boundary condition are
\begin{align}
&\left(\tilde{a} s' \right)'+\alpha s=0 \qquad&&{\rm if}\quad K\le x < p\\
&\left(\tilde{a} s' -1 \right)'=0 \qquad&&{\rm if}\quad p \le x \le p+ h \\
&\left(\tilde{a}  s' \right)'+\alpha s=0 \qquad&&{\rm if}\quad p+ h  < x \le L\\
&s= 0 \qquad && {\rm on} \quad x=K \label{bc:1d1-1}\\
&s = 0 \qquad && {\rm on} \quad x=L \label{bc:1d1-2}
\end{align}
The analytical solution is derived as follows:
\begin{equation}
s(x) =
\begin{cases}
c_1 e^{\lambda x} +c_2 e^{-\lambda x}\qquad&{\rm if}\quad K\le x < p\\
c_3 x+c_4 \qquad&{\rm if}\quad p \le x \le p+ h \\
c_5 e^{\lambda x} +c_6 e^{-\lambda x} \qquad&{\rm if}\quad p + h  < x \le L
\end{cases}
\end{equation}
where $\lambda$ is $\lambda=\sqrt{\alpha/\tilde{a}}$. 
In addition, the thickness function $h_f(s)$ is expressed as follows:
\begin{equation}
h_f\left( s \right)=
\left(
\frac{1}{h_0}
\frac{1}{c_3}
-h_{0} a
\right) \bm{1}_\Omega
\end{equation}
The constants $c_i$ are determined based on the boundary conditions (\ref{bc:1d1-1}) and (\ref{bc:1d1-2}), as well as the continuous conditions with respect to the state variable $s$ and normal flux at $x=p$ and $x=p+ h $:
\begin{equation*}
c_1=-\tfrac{h}{\tilde{a}}
\left(\tfrac{   e^{\lambda(2L+p)} -e^{\lambda(3p+2 h )} }
{
	(2+\lambda  h ) e^{2\lambda (L+p)}
	+(\lambda  h  e^{2\lambda L} +(2-\lambda  h )e^{2\lambda (p+ h )})e^{2\lambda K}
	+\lambda  h  e^{2 \lambda (2p + h )}
} \right)
\end{equation*}
\begin{equation*}
c_2=\tfrac{h e^{2\lambda K}}{\tilde{a}}
\left(\tfrac{   e^{\lambda(2L+p)} +e^{\lambda(3p+2 h )} }
{
	(2+\lambda  h ) e^{2\lambda (L+p)}
	+(\lambda  h  e^{2\lambda L} +(2-\lambda  h )e^{2\lambda (p+ h )})e^{2\lambda K}
	+\lambda  h  e^{2 \lambda (2p + h )}
} \right)
\end{equation*}
\begin{equation*}
c_3=-\tfrac{2}{\tilde{a}}
\left(\tfrac{   e^{2\lambda(K+p+ h )} -e^{2\lambda(L+p)} }
{
	(2+\lambda  h ) e^{2\lambda (L+p)}
	+(\lambda  h  e^{2\lambda L} +(2-\lambda  h )e^{2\lambda (p+ h )})e^{2\lambda K}
	+\lambda  h  e^{2 \lambda (2p + h )}
} \right)
\end{equation*}
\begin{equation*}
c_4=\tfrac{1}{\tilde{a}}
\left(\tfrac{ 
	-(2p+ h )e^{2\lambda (L+p)}
	+(h e^{2\lambda L} +(2p+ h ) e^{2\lambda (p+ h )}) e^{2\lambda K}
	-h e^{2\lambda  (2p+ h )}
}
{
	(2+\lambda  h ) e^{2\lambda (L+p)}
	+(\lambda  h  e^{2\lambda L} +(2-\lambda  h )e^{2\lambda (p+ h )})e^{2\lambda K}
	+\lambda  h  e^{2 \lambda (2p + h )}
} \right)
\end{equation*}
\begin{equation*}
c_5 =-\tfrac{h}{\tilde{a}}
\left(\tfrac{   e^{\lambda(2K+p+ h )} + e^{\lambda(2p+ h )} }
{
	(2+\lambda  h ) e^{2\lambda (L+p)}
	+(\lambda  h  e^{2\lambda L} +(2-\lambda  h )e^{2\lambda (p+ h )})e^{2\lambda K}
	+\lambda  h  e^{2 \lambda (2p + h )}
} \right)
\end{equation*}
\begin{equation*}
c_6 =\tfrac{h}{\tilde{a}}
\left(\tfrac{ \displaystyle  e^{\lambda(2L+2K+p+ h )} + e^{\lambda(2L+3p+ h )} }
{
	(2+\lambda  h ) e^{2\lambda (L+p)}
	+(\lambda  h  e^{2\lambda L} +(2-\lambda  h )e^{2\lambda (p+ h )})e^{2\lambda K}
	+\lambda  h  e^{2 \lambda (2p + h )}
} \right)
\end{equation*}
If the black domain is sufficiently far from the boundaries $x=K$ and $x=L$, the coefficients are simplified as follows:
\begin{align*}
&
\lim_{\substack{K\rightarrow -\infty \\ L\rightarrow \infty} } c_1=
-\frac{h e^{-\lambda p}}{\tilde{a}(\lambda h+2)}
&&
\lim_{\substack{K\rightarrow -\infty \\ L\rightarrow \infty} } c_2 = 0
&&&
\lim_{\substack{K\rightarrow -\infty \\ L\rightarrow \infty} } c_3 =
\frac{2}{\tilde{a}(\lambda h+2)}\\[2pt]
&
\lim_{\substack{K\rightarrow -\infty \\ L\rightarrow \infty} } c_4=
-\frac{2p + h}{\tilde{a}(\lambda h+2)}
&&
\lim_{\substack{K\rightarrow -\infty \\ L\rightarrow \infty} } c_5= 0
&&&
\lim_{\substack{K\rightarrow -\infty \\ L\rightarrow \infty} } c_6=
\frac{h e^{\lambda (p+h)}}{\tilde{a}(\lambda h+2)}
\end{align*}
Therefore, the state variable $s$ is simplified as follows:
\begin{equation}
\lim_{\substack{K\rightarrow -\infty \\ L\rightarrow \infty} }  
s(x)=
\begin{cases}
\displaystyle
\frac{h}{\tilde{a}(\lambda h+2)}e^{\lambda(x-p)} \qquad &{\rm if}\quad  x < p\\[2pt]
\displaystyle
\frac{1}{\tilde{a}(\lambda h+2)} \Bigl( 2x-(2p+h)\Bigl)
\qquad &{\rm if}\quad  p \le x \le p+h\\[2pt]
\displaystyle
\frac{h}{\tilde{a}(\lambda h+2)}e^{-\lambda(x-(p+h))} \qquad &{\rm if}\quad  p+h < x.
\end{cases}
\label{eq:1d}
\end{equation}
Note that the role of the black domain is to describe the damping of the state variable. 
Therefore, the order of sufficient distance from the black domain is $\tfrac{1}{\lambda}$, which represents the sharp area of the function $e^{-\lambda x}$.

Finally, the thickness function is obtained as follows:
\begin{align}
\lim_{\substack{K\rightarrow -\infty \\ L\rightarrow \infty} }
h_f\left(s \right)&=
\begin{cases}
0 \qquad &{\rm if}\quad  x < p\\[2pt]
h&{\rm if}\quad  p \le x \le p+h\\[2pt]
0 \qquad &{\rm if}\quad  p+h < x 
\end{cases}.
\end{align}
The value of the thickness function in the black domain is exactly equivalent to its thickness $h$, as long as the black domain is sufficiently far from the boundaries of the reference domain. To satisfy this condition, the damping coefficient are set to a relatively large value.
If the black domain is relatively far from another black domain, the thickness function in the black domain is also exactly equivalent to its thickness $h$. This case is also verified by considering the periodic domain and taking the limit with respect to the period.
Here, the most notable point is that the thickness is extracted without calculating a distance.

The orientation function is well defined because one-dimensional orientation is a sign of the state variable $s$. 
The skeleton is defined along a dimension without a base transformation. Then, the skeleton clearly indicates a point at the center of the black domain because the distribution of the state variable $s(x)$ described in Equation (\ref{eq:1d}) is zero at $x=p+\frac{h}{2}$.

The analytical solution of the one-dimensional case can now be presented. 
Cases (a), (b), and (c) show the effects on $h$, $p$, and $a$, respectively.
These parameters are listed in Table \ref{tab:1d}.
\begin{table}[htb]
	\begin{center}
		\caption{Parameters in one-dimensional calculations}
		\begin{tabular}{ccccccc}\hline
			& $K$  	&  $L$ 	 & $h_0$& $h$& $p$& $a$ \\ \hline
			case 1  	 & $0.0$ & $1.0$ & $0.2$& -- & $0.2$& $0.2$  \\
			case 2  	 & $0.0$ & $1.0$ & $0.2$& $0.2$ & --& $0.2$  \\
			case 3  	 & $0.0$ & $1.0$ & $0.2$& $0.2$ & $0.4$& --  \\ \hline
			\label{tab:1d}
		\end{tabular}
	\end{center}
\end{table}
\begin{figure*}[htb]
	\begin{center}
		\subfigure[ effect on $h$]{
			\includegraphics[width=4.1cm]{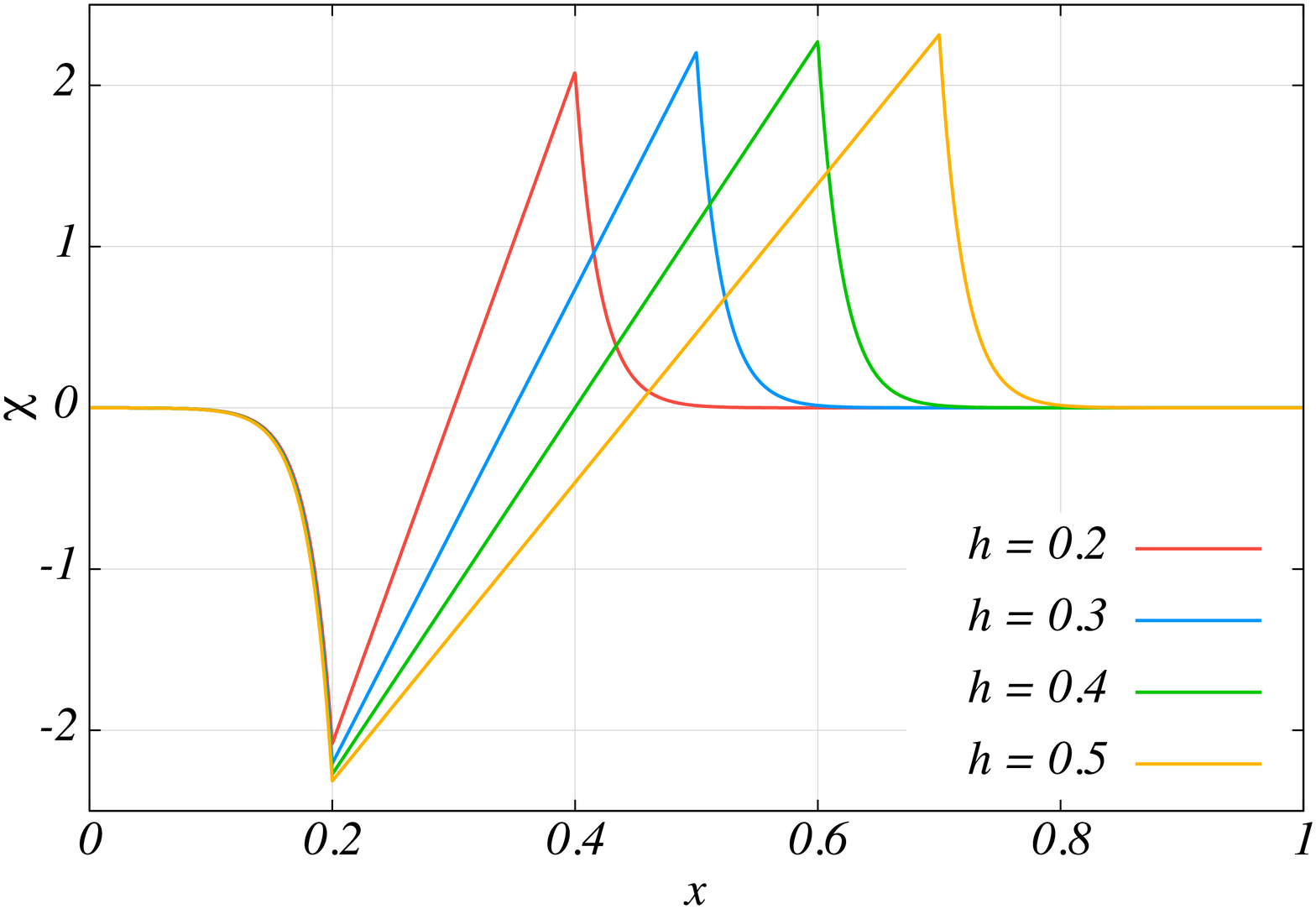}}
		\subfigure[ effect on $p$]{
			\includegraphics[width=4.1cm]{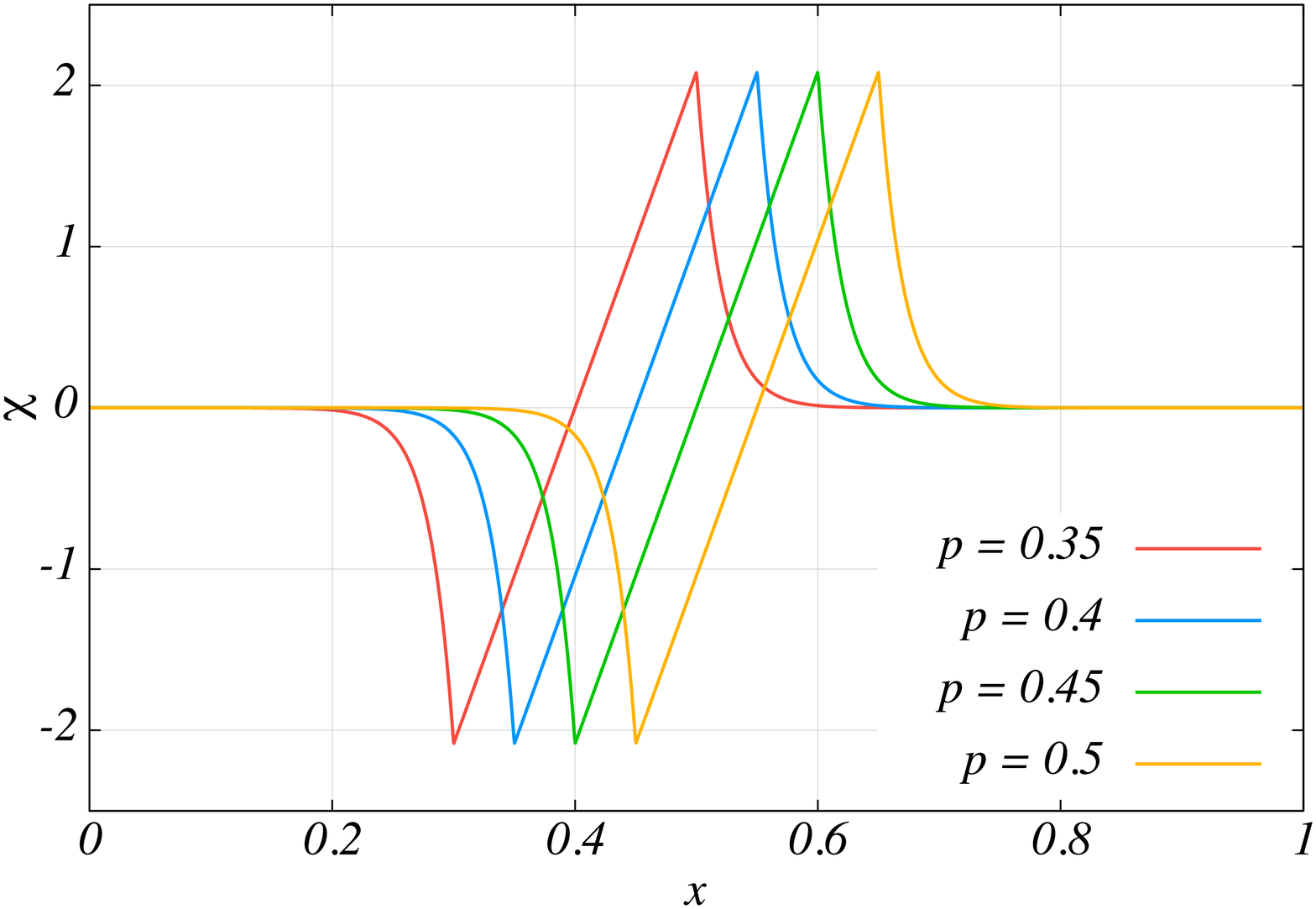}}
		\subfigure[ effect on $a$ ]{
			\includegraphics[width=4.1cm]{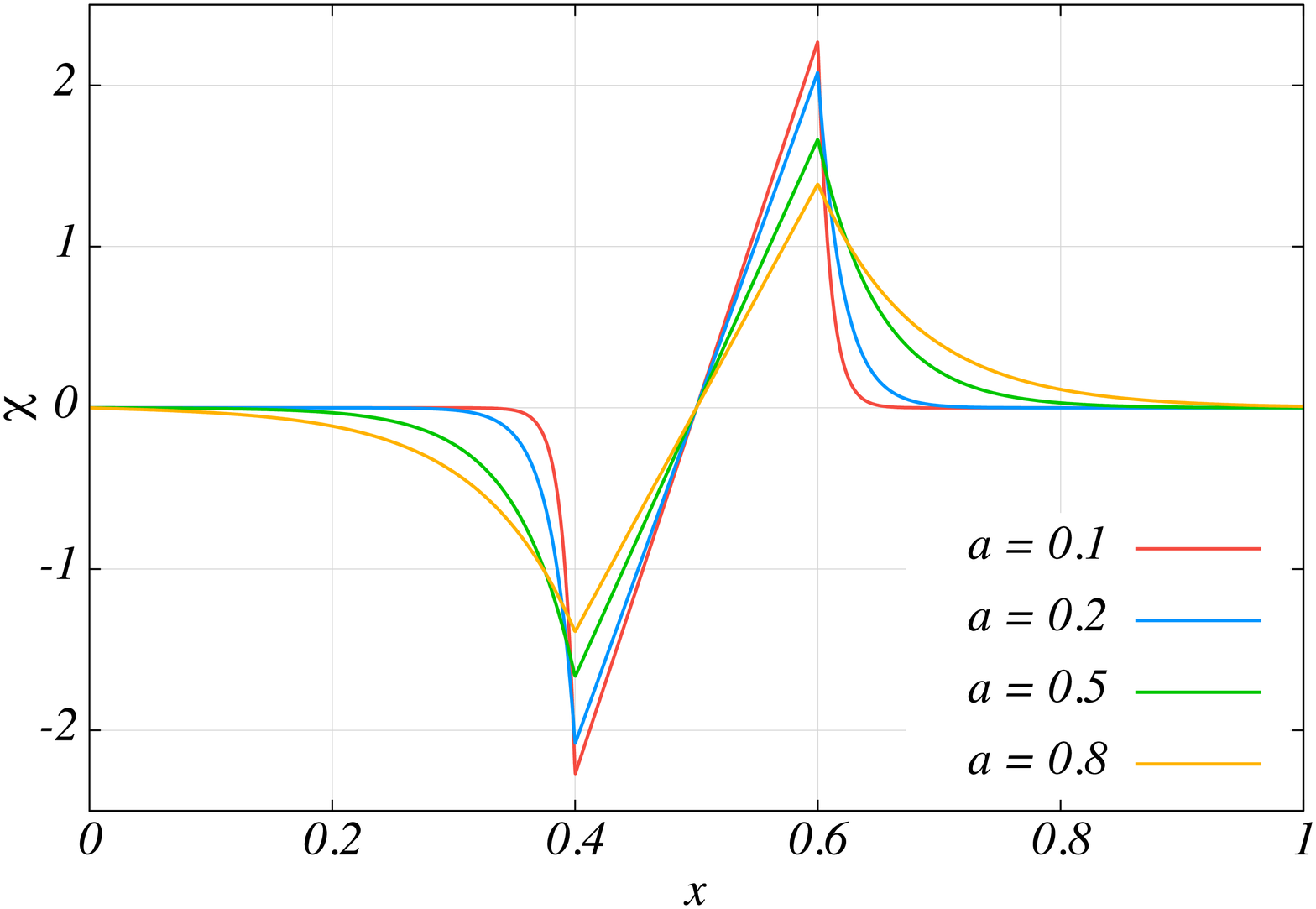}}
		\caption{Analytical solutions of the one-dimensional case}
		\label{fig:1dResult}
	\end{center}
\end{figure*}
As shown in Figure \ref{fig:1dResult}, the solution $s$ is exponentially damped in the white domain and is a linear function in the black domain $\Omega$. 
The profile of the state variable $s$ is fixed when the parameter $p$ varies as shown in Figure \ref{fig:1dResult}(b). Therefore, a shape is equivalently evaluated to the shape whose position has changed.
In addition, the effect from another shape or boundary is neglected when the damping coefficient is set to a relatively large value. This is because the distribution of the state variable $s$ converges exponentially to zero in the white domain, as shown in Figure \ref{fig:1dResult}(c).
Therefore, it is confirmed that the proposed method can be used to correctly evaluate the geometric features in one dimension.

The characteristic length $h_0$ should be set to approximately the smallest target thickness, because the proposed method is not satisfied in a relatively small shape owing to the diffusive nature of the PDE.
In other words, relatively small shape fractionation is neglected because the local information in the PDE system is averaged by the diffusion term.
Although a smaller diffusion coefficient provides a more precise evaluation, a small diffusion coefficient is required for a fine finite element mesh in numerical computation. Therefore, the value of $a$ must satisfy the aforementioned condition. In addition, the finite element mesh size is determined by the value of parameter $a$.  
%------------------------------------------------------------------
%--------------------------------------------------------------------------------------
% Section 
%--------------------------------------------------------------------------------------
%--------------------------------------------------------------------------------------
\section{Numerical validation in two dimensions}\label{sec:2d}
\subsection{Multiple bars}
\begin{figure*}[htb]
	\begin{center}
		\subfigure[Input image data, where the image size is $5\times 8$. The black and white colors show the black and white domains, respectively.]{
			\includegraphics[height=2cm]{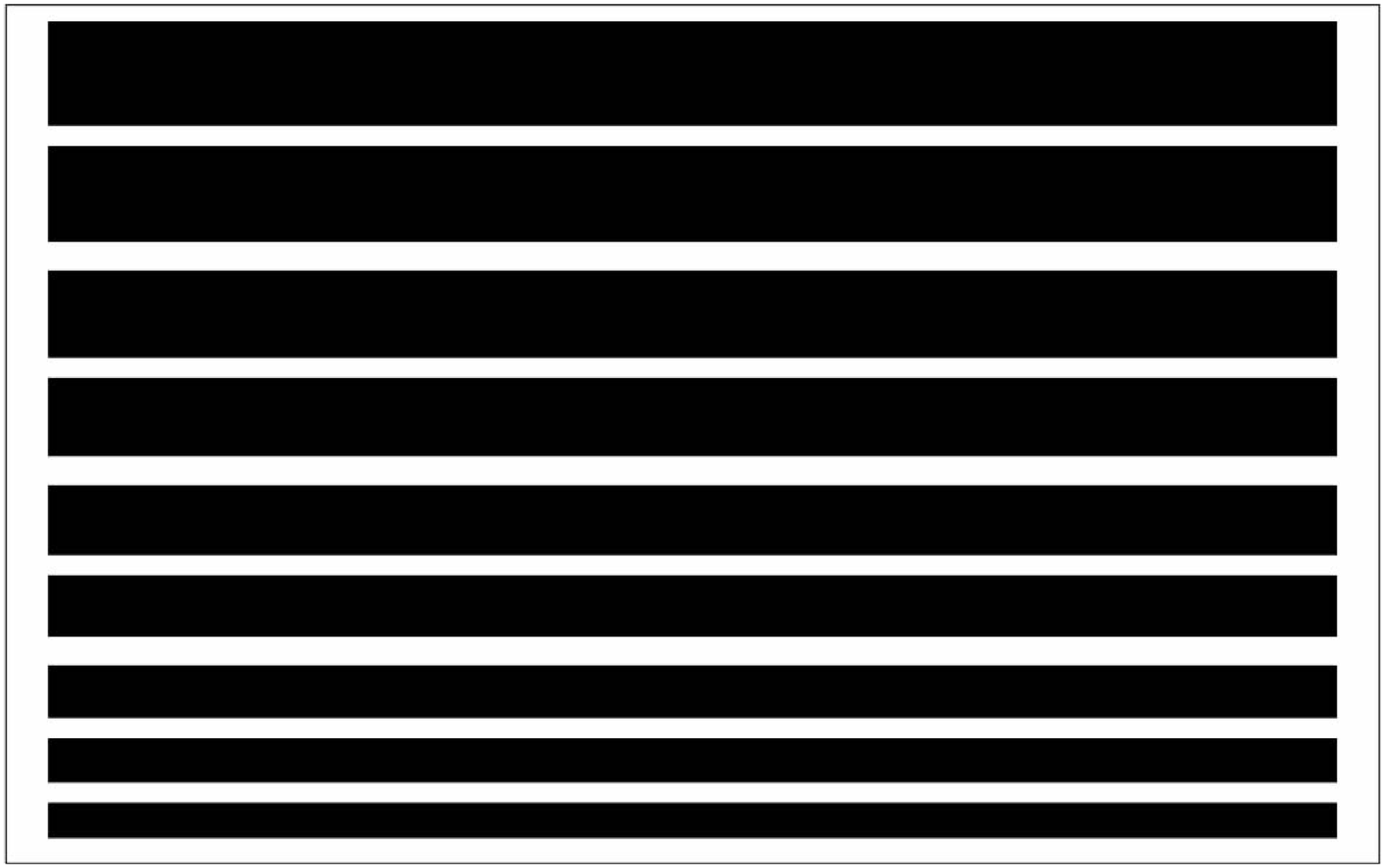}}\quad
		\subfigure[Orientation vectors, where the blue and red colors indicate the normal and tangential vectors, respectively. The gray and white domain colors are the input image data for reference.]{
			\includegraphics[height=2cm]{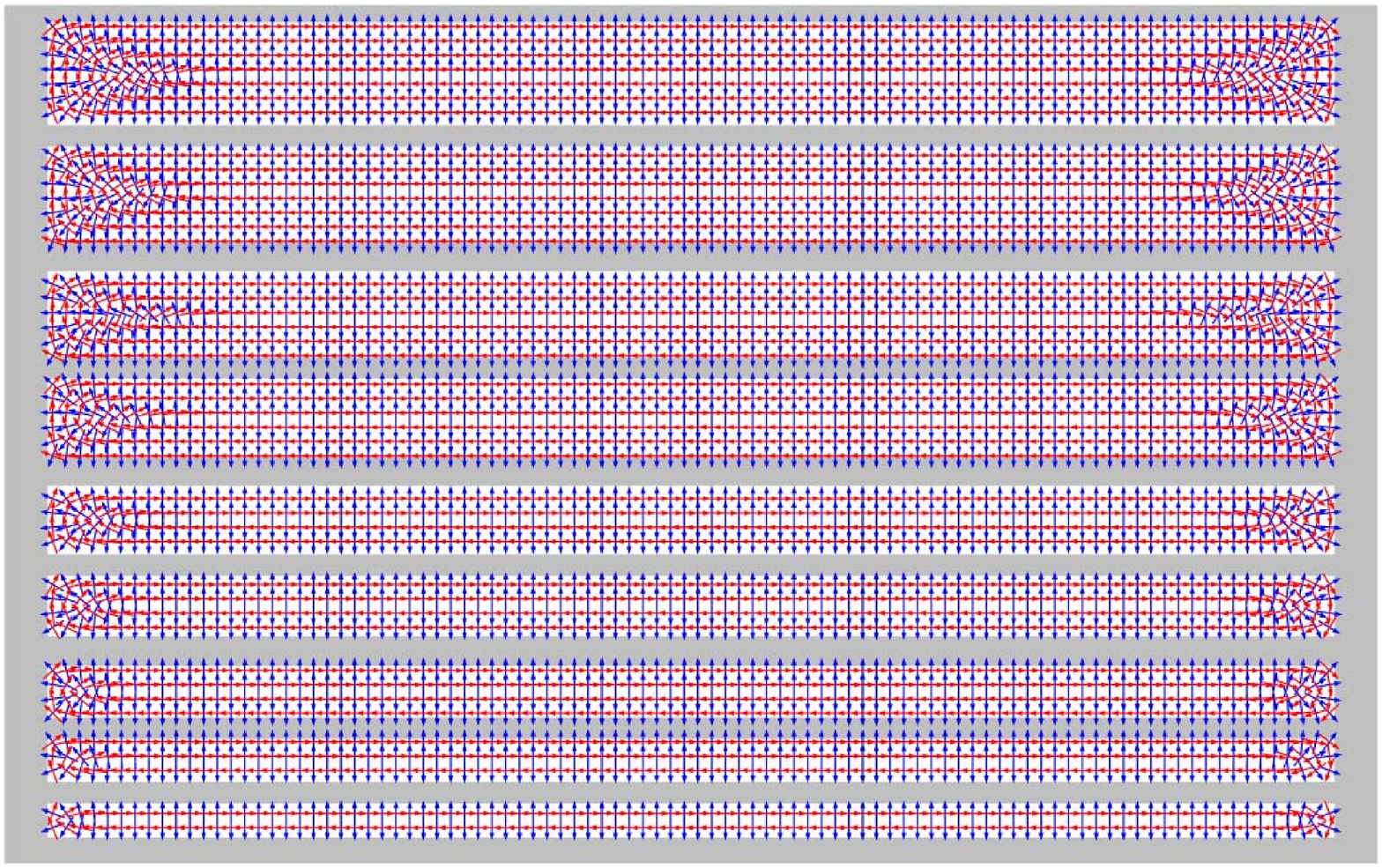}}\quad
		\subfigure[Skeleton function $f_s (s_1, s_2)$. The gray and white domain colors are the input image data for reference.]{
			\includegraphics[height=2cm]{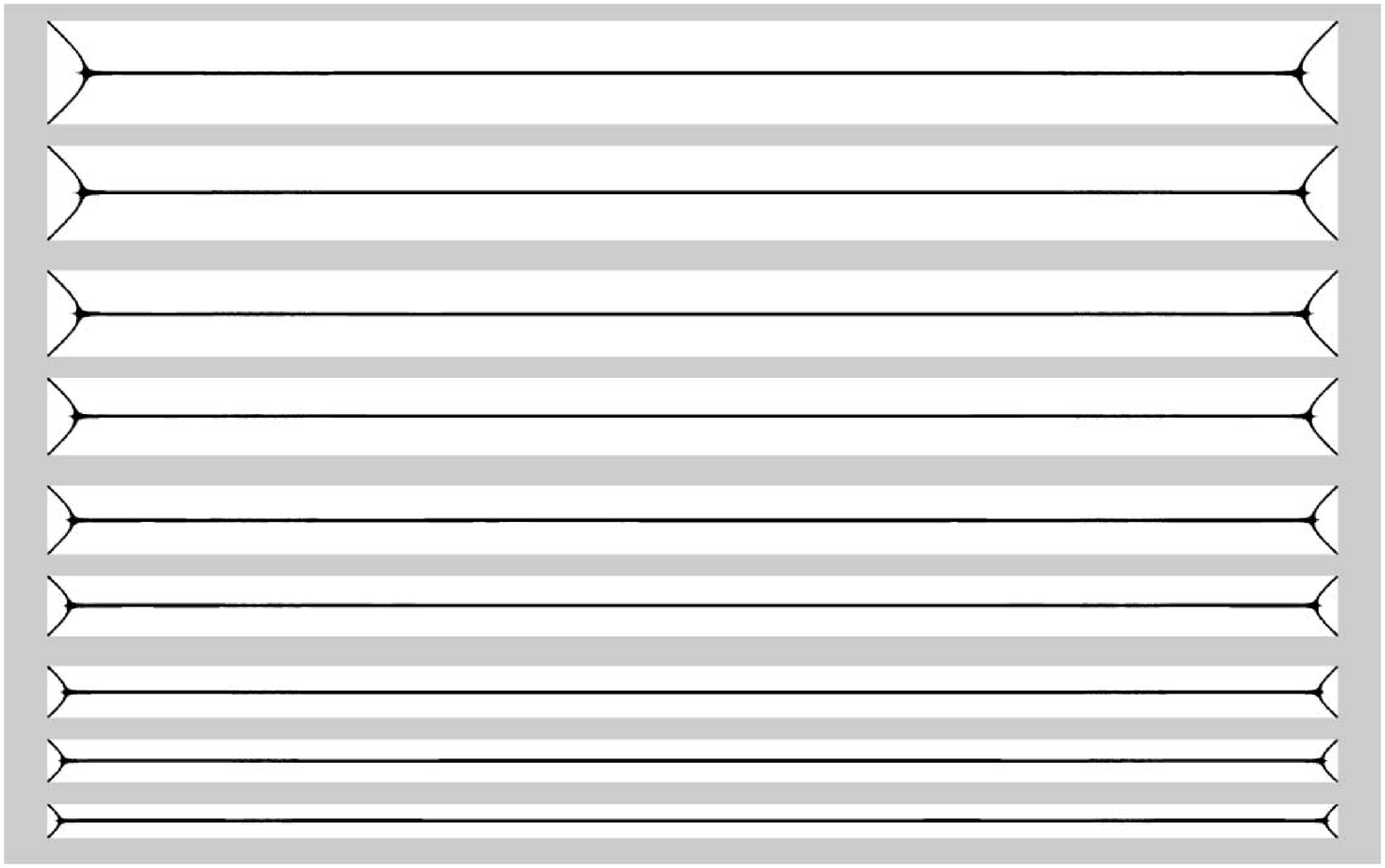}}\\
		\subfigure[Distribution of the state variable $s_1$]{
			\includegraphics[height=2cm]{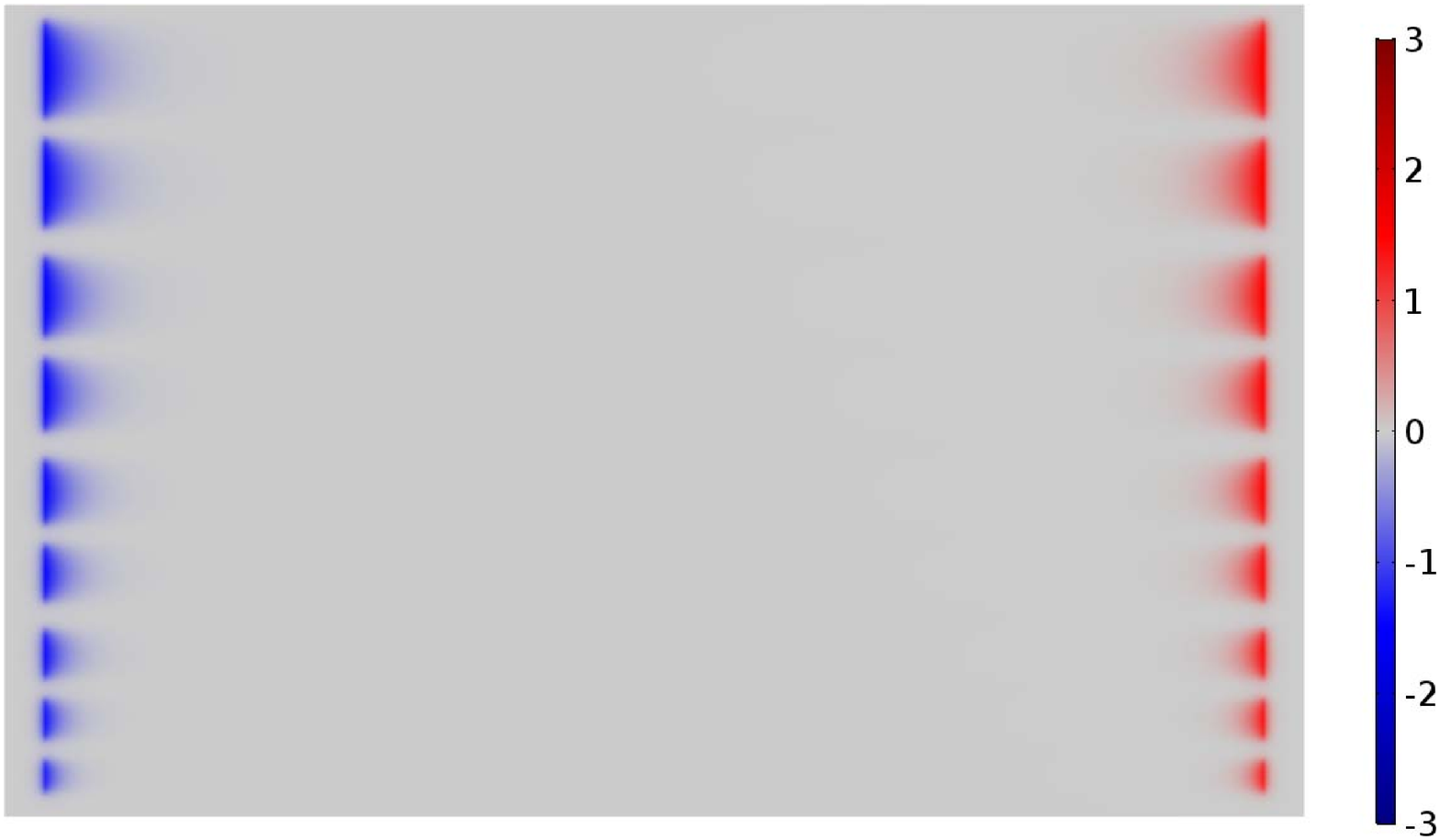}}
		\subfigure[Distribution of the state variable $s_2$ ]{
			\includegraphics[height=2cm]{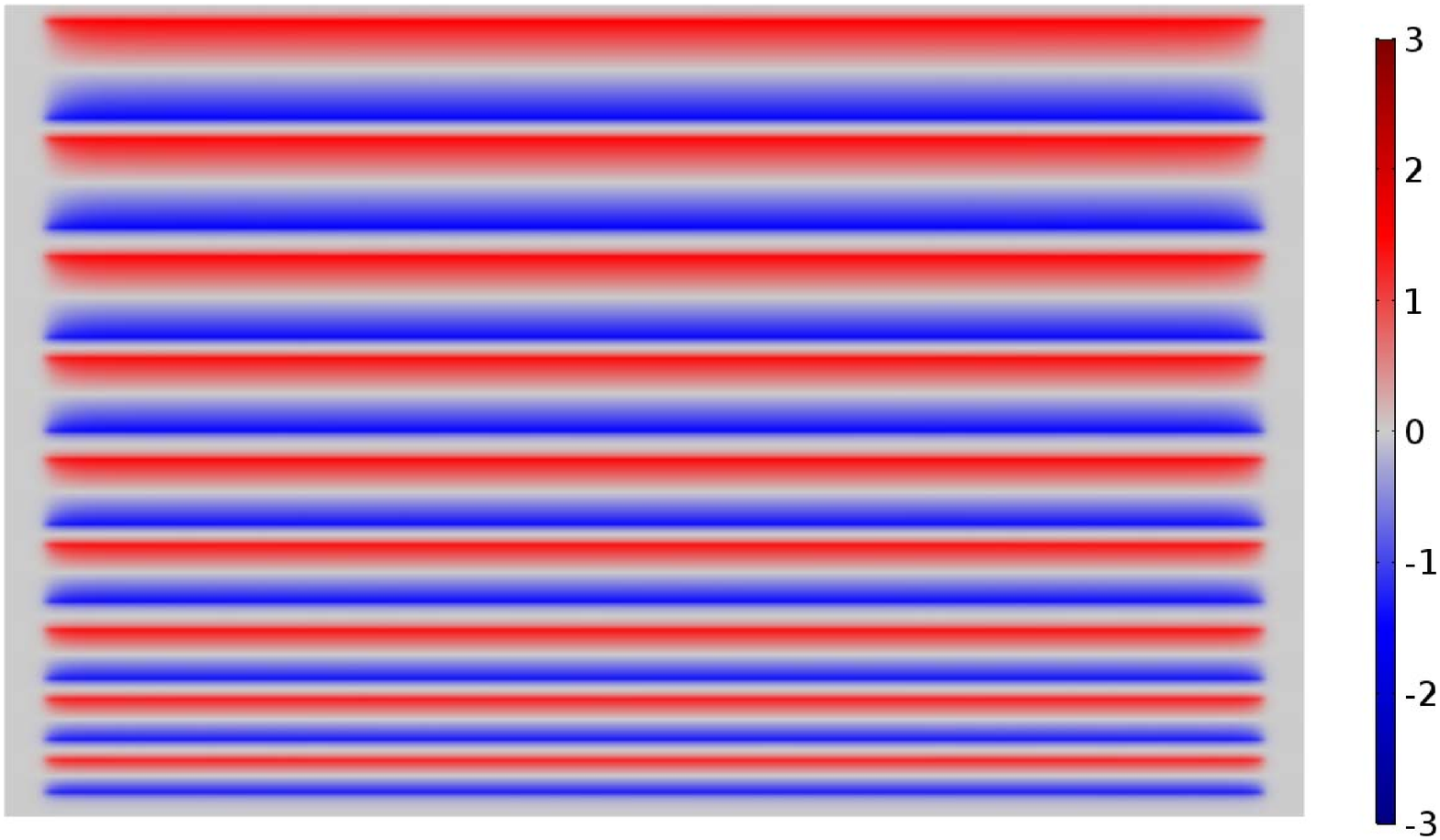}}
		\subfigure[Thickness function $h_f (s_1, s_2)$]{
			\includegraphics[height=2cm]{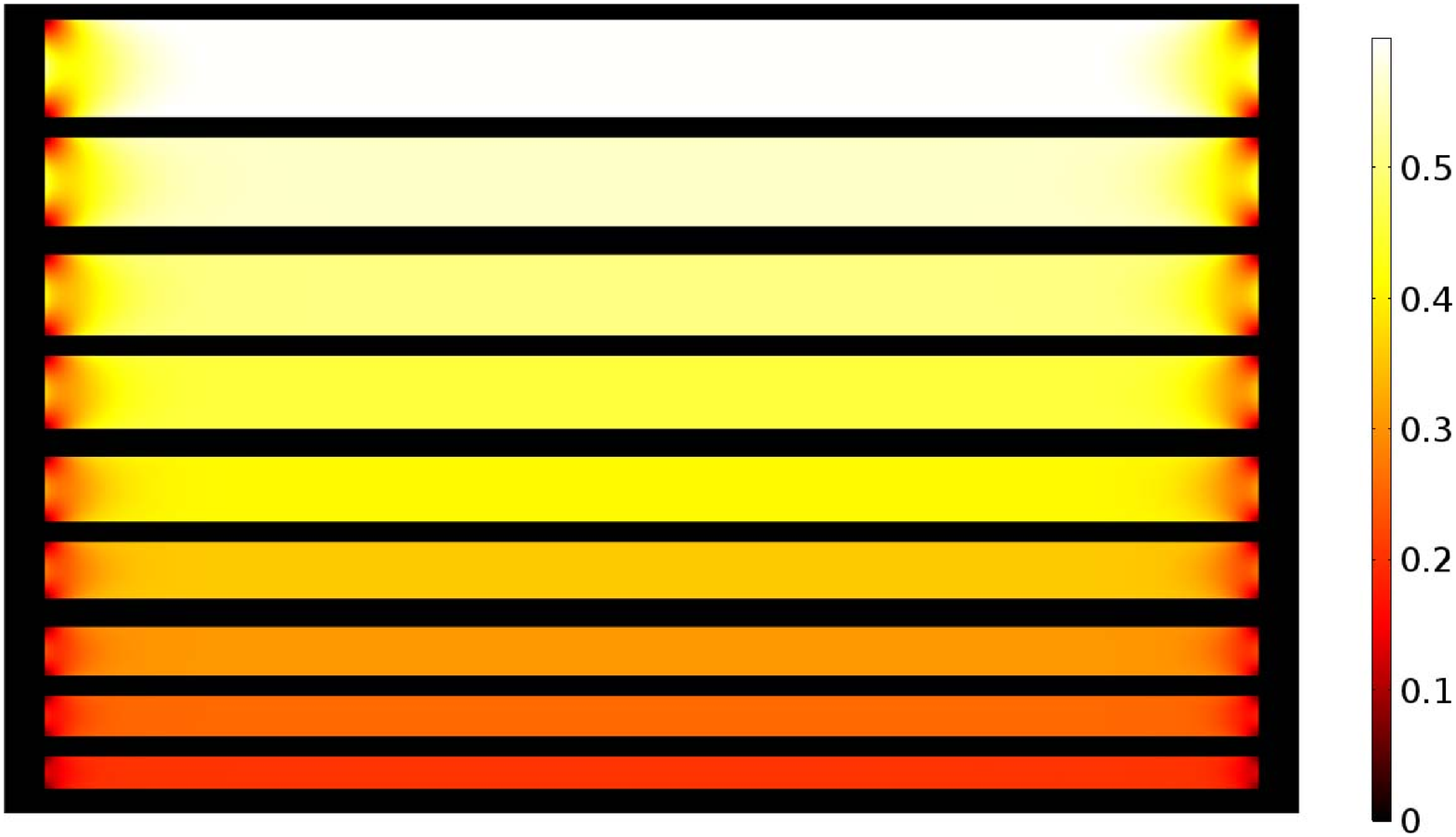}}\\
		\subfigure[Inverse thickness function $f_h (s_1, s_2)$]{
			\includegraphics[height=2cm]{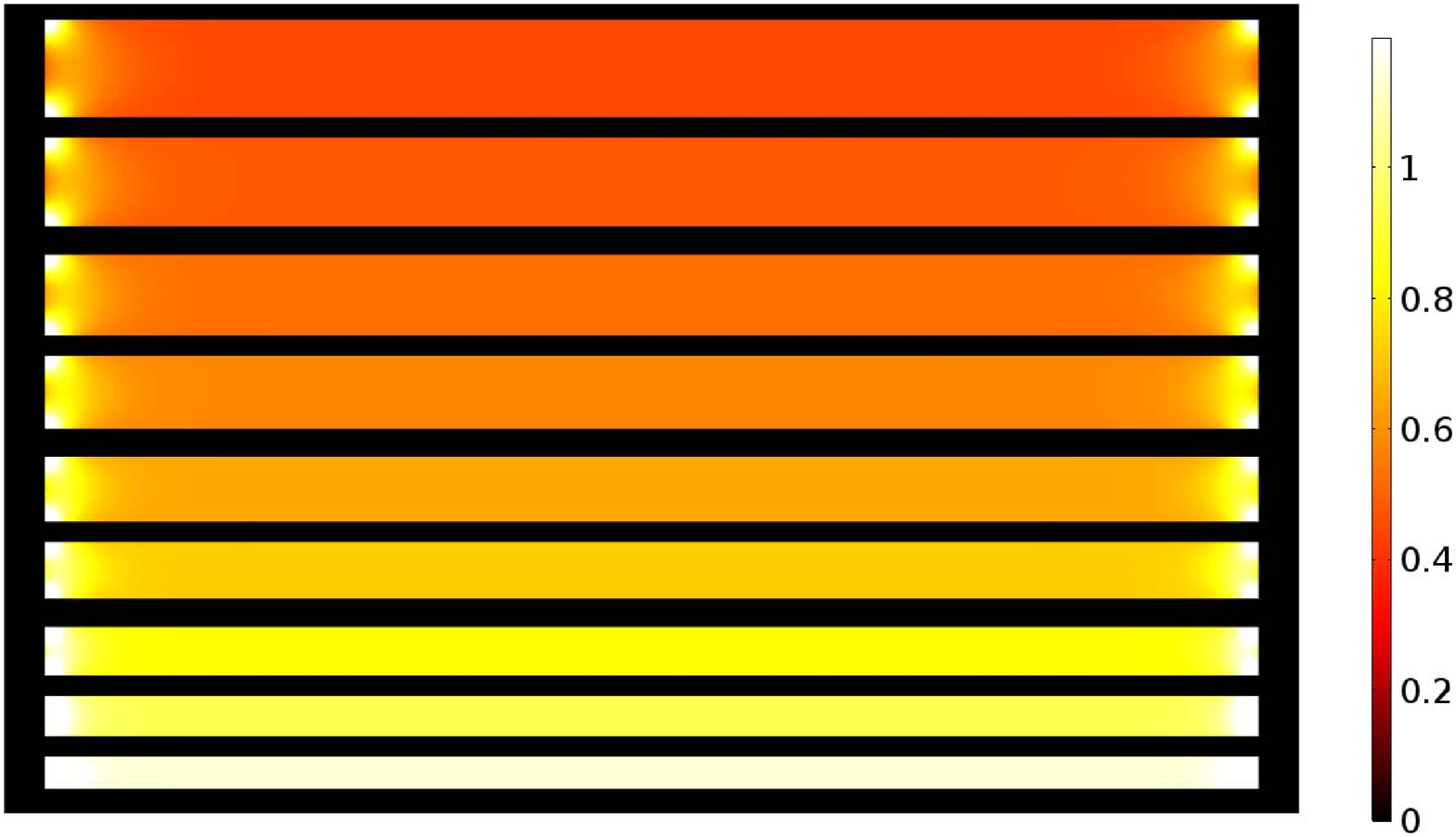}}
		\qquad
		\subfigure[Plot of the average value of $\frac{1}{f_h}$ in the central domain of each bar and the linear function estimated using the least-squares method]{
			\includegraphics[height=3cm]{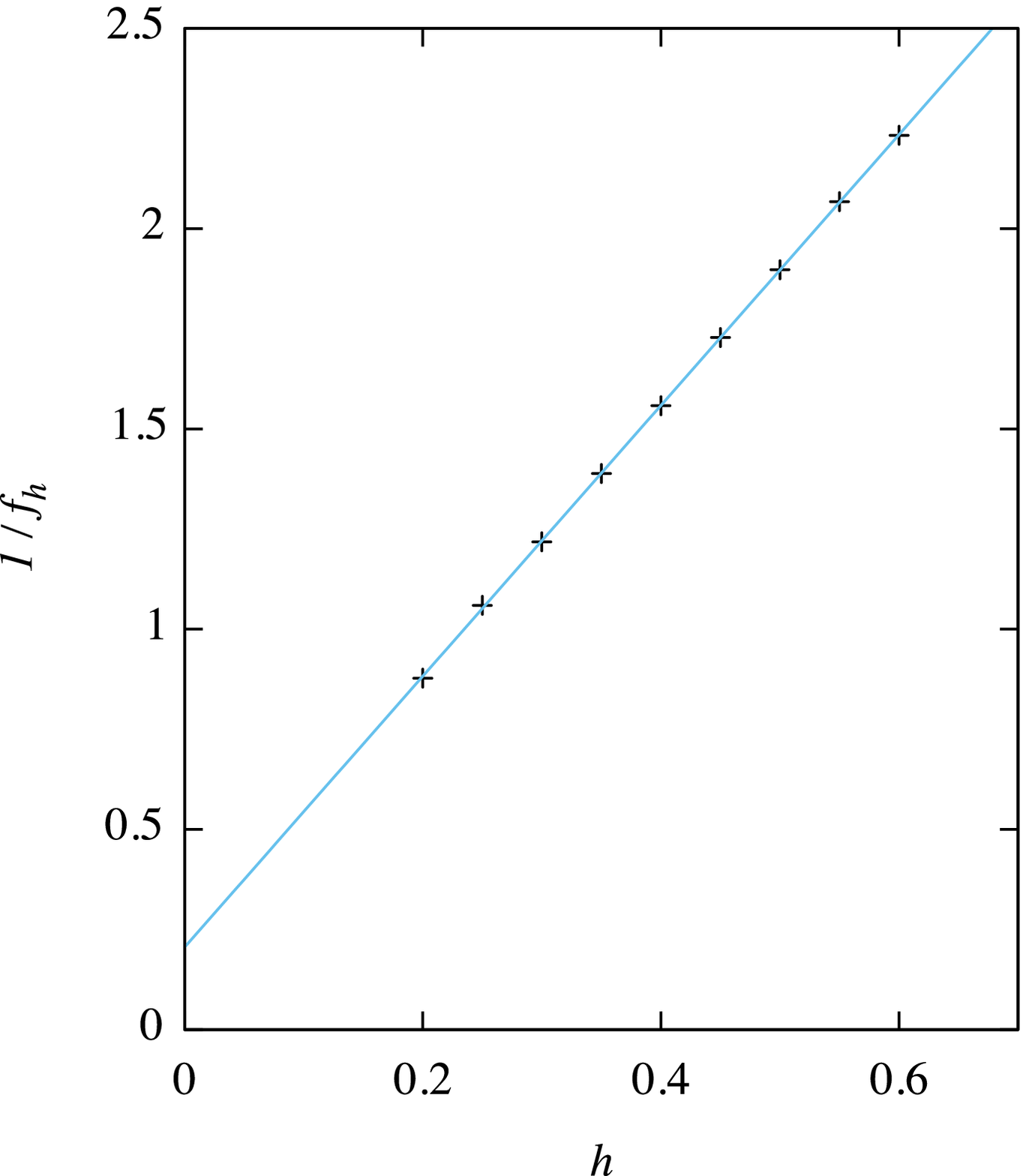}}\quad
		\subfigure[Plot of the average value of $h_f$ in the central domain of each bar and the linear function estimated using the least-squares method]{
			\includegraphics[height=3cm]{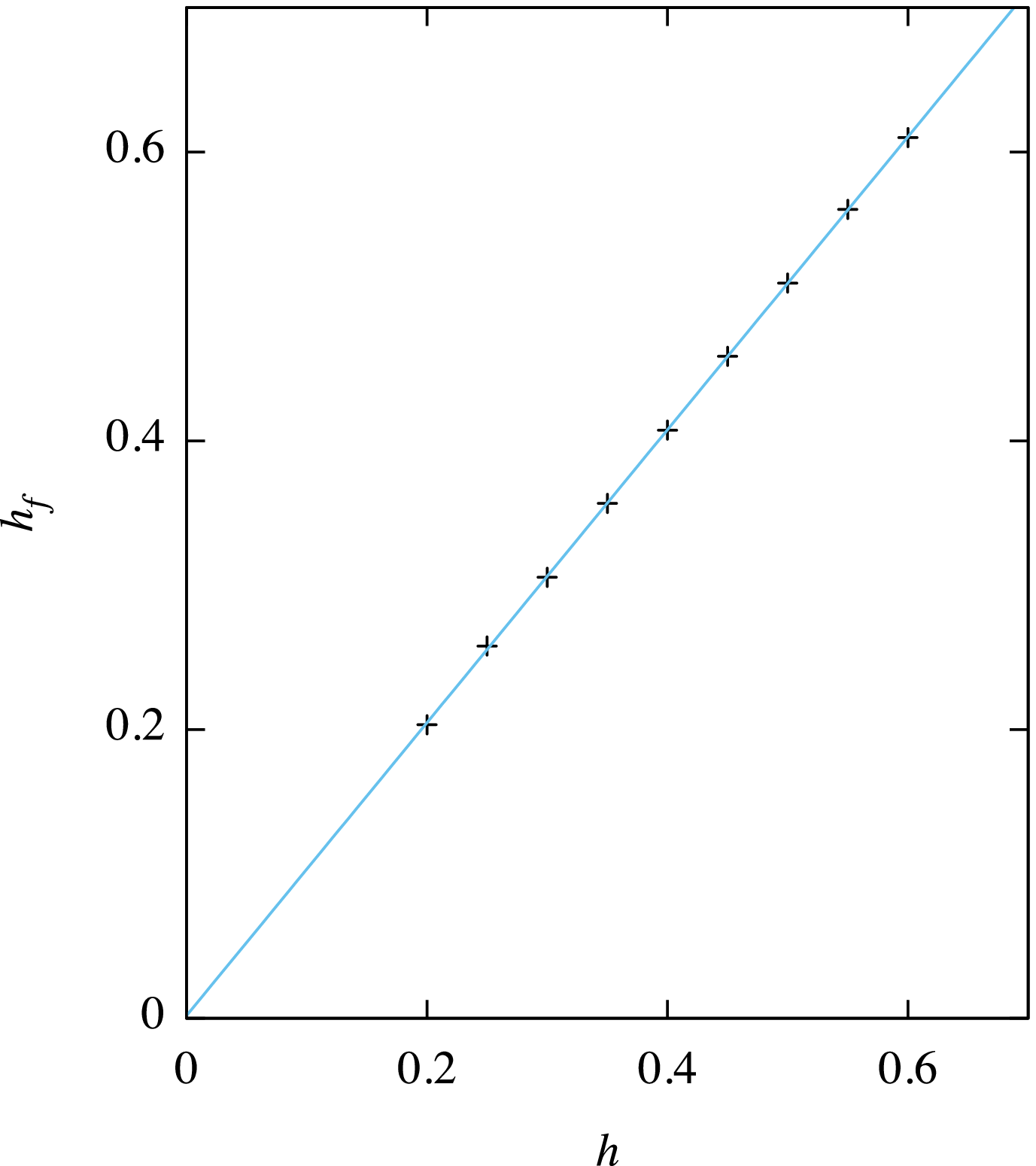}}
		\caption{Numerical results for an image of multiple bars with different thicknesses}
		\label{fig:2dResult1}
	\end{center}
\end{figure*}

The two-dimensional case shown in Figure \ref{fig:2dResult1} (a) is considered here.
As shown in the figure, the input image data includes multiple bars with thicknesses $h$ of $0.2$, $0.25$, $0.3$, $0.35$, $0.4$, $0.45$, $0.55$, and $0.6$. The image size is $5\times 8$.
The parameters of the PDE are set to $h_0=0.3$ and $a=0.2$.

The domain is discretized using $\PP 2$ triangular finite elements with the maximum length of $0.05$. 
The obtained state variables $s_1$ and $s_2$ are shown in Figures \ref{fig:2dResult1} (d) and (e).
The normal orientation vector  $\bm{n}_f(s_1, s_2)$, tangential orientation vector $\bm{t}_f(s_1, s_2)$, inverse thickness function $f_h(s_1, s_2)$, thickness function $h_f(s_1, s_2)$, and skeleton function $f_s(s_1, s_2)$ are shown in Figure \ref{fig:2dResult1}.

The properties of the inverse thickness function $f_h$ are discussed first.
As shown in Figure \ref{fig:2dResult1}(g), the values are constant in the black domain, excluding the corners.
The relationship between each value of the inverse thickness function $f_h$ and the thickness of the bar is plotted in Figure \ref{fig:2dResult1}(h), where the longitude and abscissa axes are the average value of $1/f_h$ and the thickness of the bars, respectively. The average values are computed in each central domain where the width is $5$ to avoid the corner effect. 
The linear function estimated with the least-squares method is shown in Figure \ref{fig:2dResult1}(h). The coefficient of determination is $R^2=1.0000$.
This confirms that the inverse thickness function $f_h$ is inversely proportional to each thickness value.
Figure \ref{fig:2dResult1}(i) also shows the relationship between the value of the thickness function $h_f$ and the thickness of each bar. The blue line shows the linear function estimated using the least-squares method. The coefficient of determination in the estimation is also $R^2=1.0000$.
These results confirm that the thickness function $f_h$ is proportional to each thickness value.

The linear function does not intersect the origin.
Although the local thickness is precisely estimated near $h_0$, the relatively small thickness is excessively estimated. This is because of the diffusion term in the PDE.
Therefore, the value of $h_0$ should be set to the smallest thickness.

The value of the thickness function $h_f$ is equivalent to each thickness value, as shown in Figure \ref{fig:2dResult1}(f). The skeleton shown in Figure \ref{fig:2dResult1}(c) is also appropriately estimated because the curve is close to the definition of a medial axis \cite{blum1967transformation}. The orientation vectors also provide good estimation, as shown in Figure \ref{fig:2dResult1}(b).
%-----------------------
\subsection{Complex shape composed of basic shapes}
The effectiveness of the proposed method in complex shapes is examined.
The image size is set to $1\times 1$. The parameters of the PDE are set to $h_0=0.3$ and $a=0.2$. The domain is discretized using $\PP 2$ triangular finite elements.

\begin{figure*}[htb]
	\begin{center}
		\subfigure[Input image data]{
			\includegraphics[height=3cm]{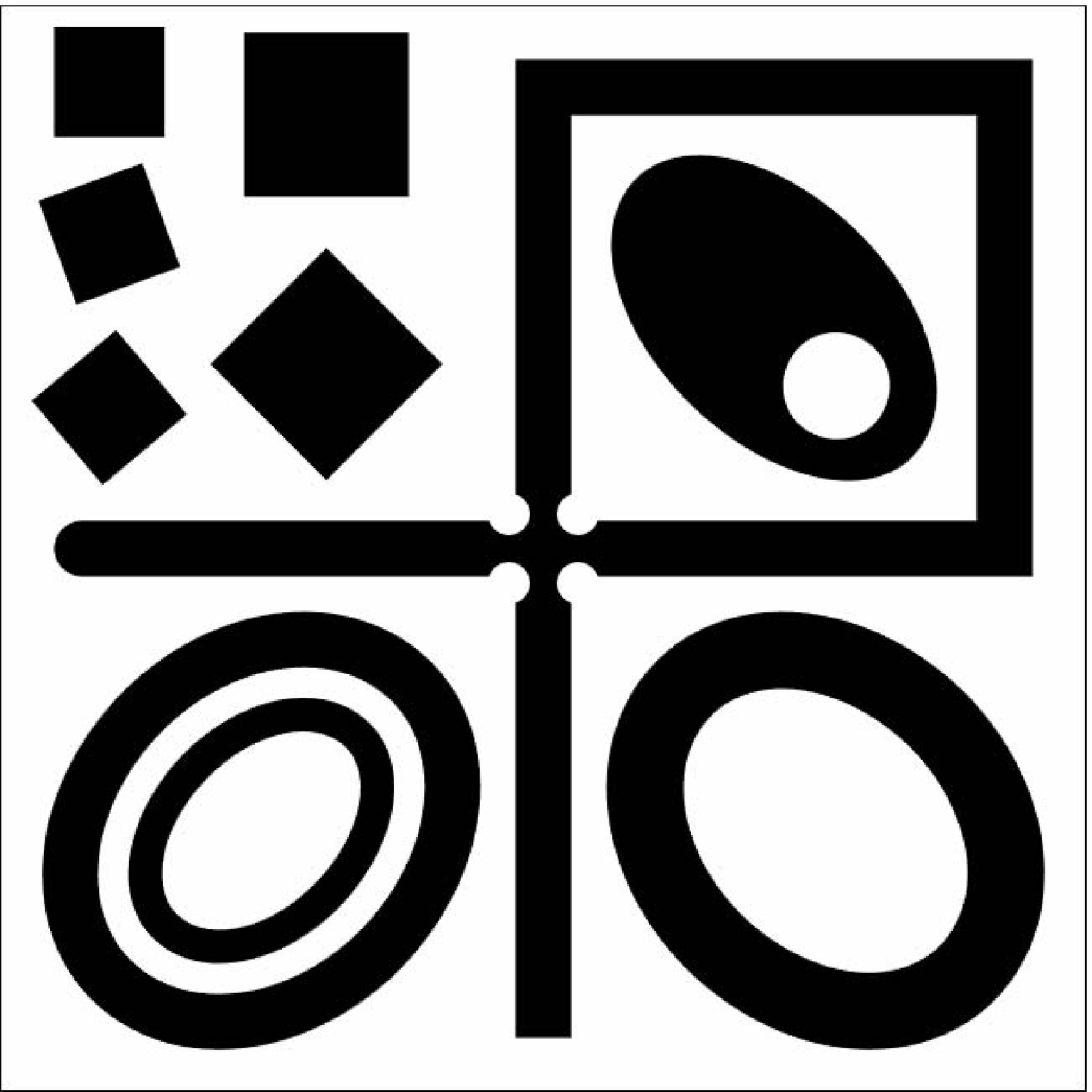}}\quad
		\subfigure[Inverse thickness function $f_h (s_1, s_2)$]{
			\includegraphics[height=3cm]{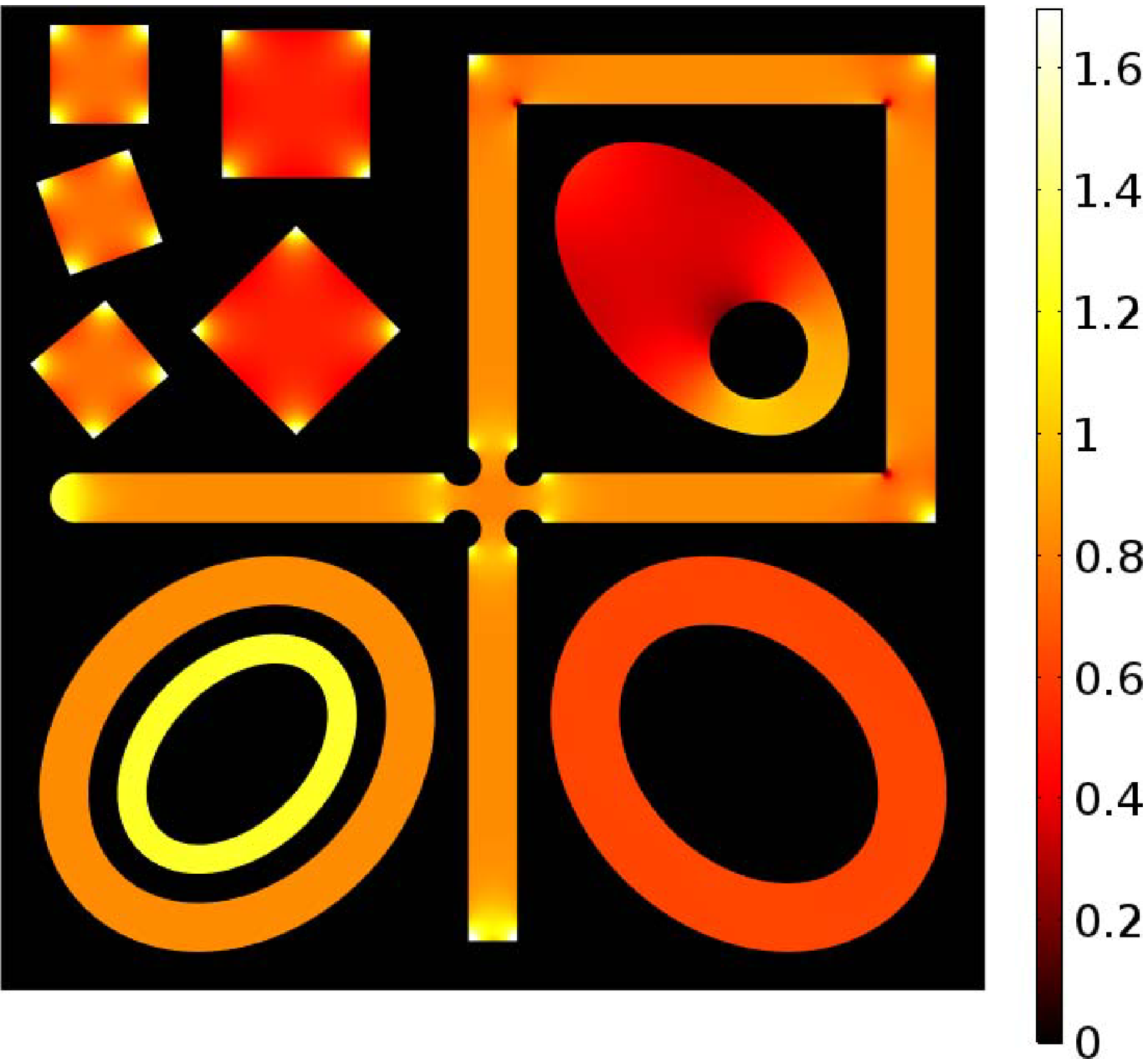}}\quad
		\subfigure[Thickness function $h_f (s_1, s_2)$]{
			\includegraphics[height=3cm]{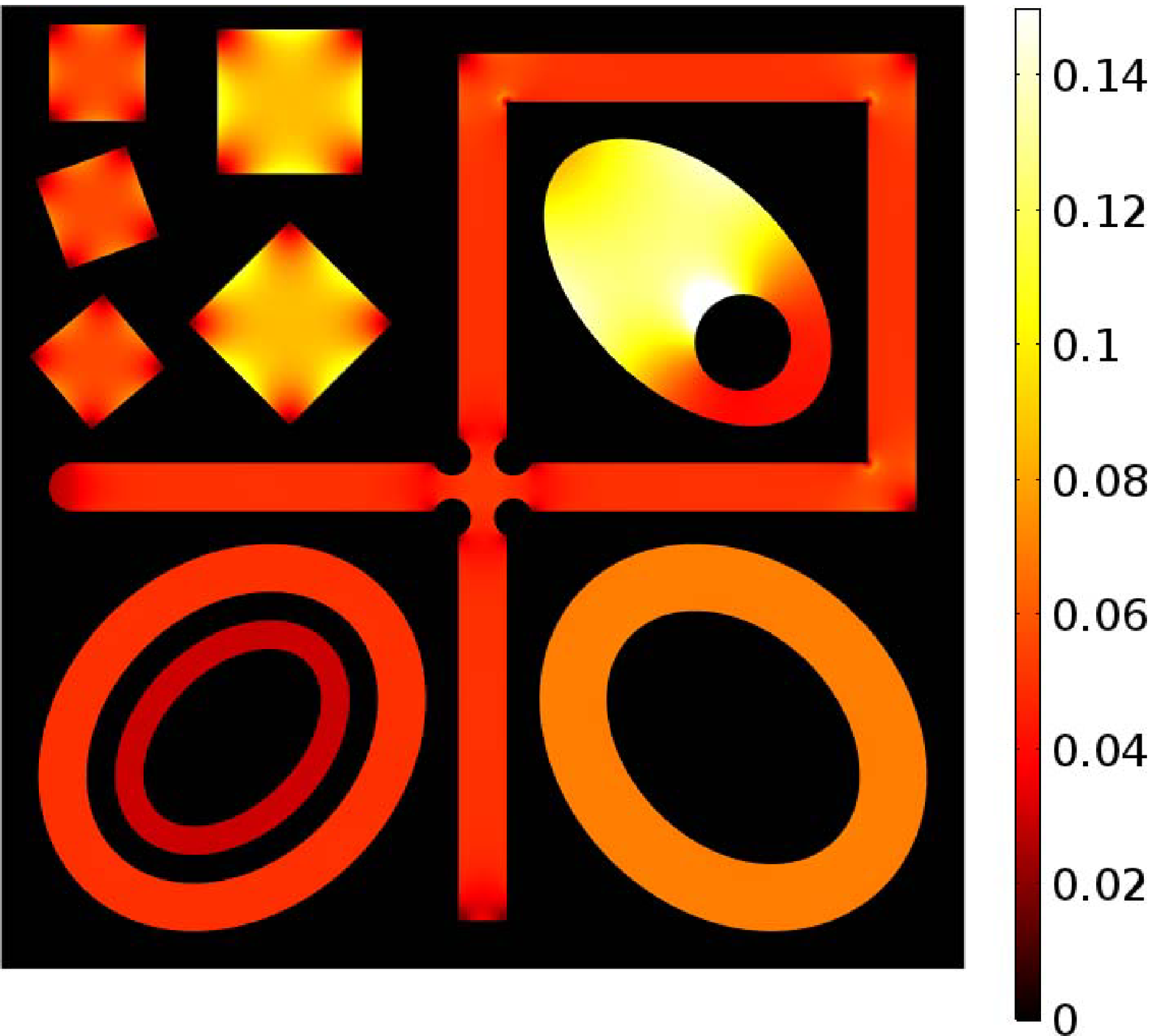}}\quad
		\subfigure[Orientation vectors]{
			\includegraphics[height=3cm]{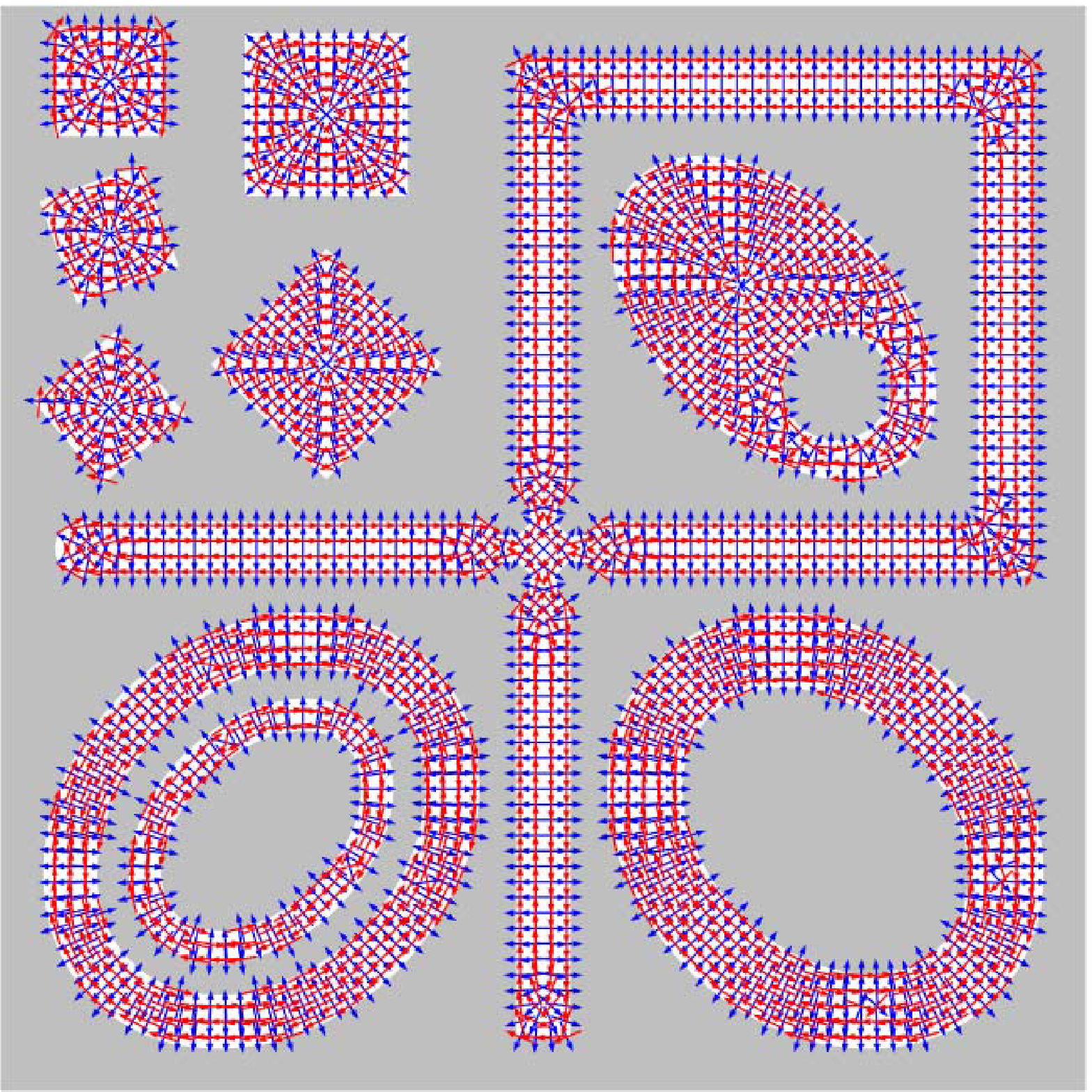}}\quad
		\subfigure[Skeleton function $f_s (s_1, s_2)$.]{
			\includegraphics[height=3cm]{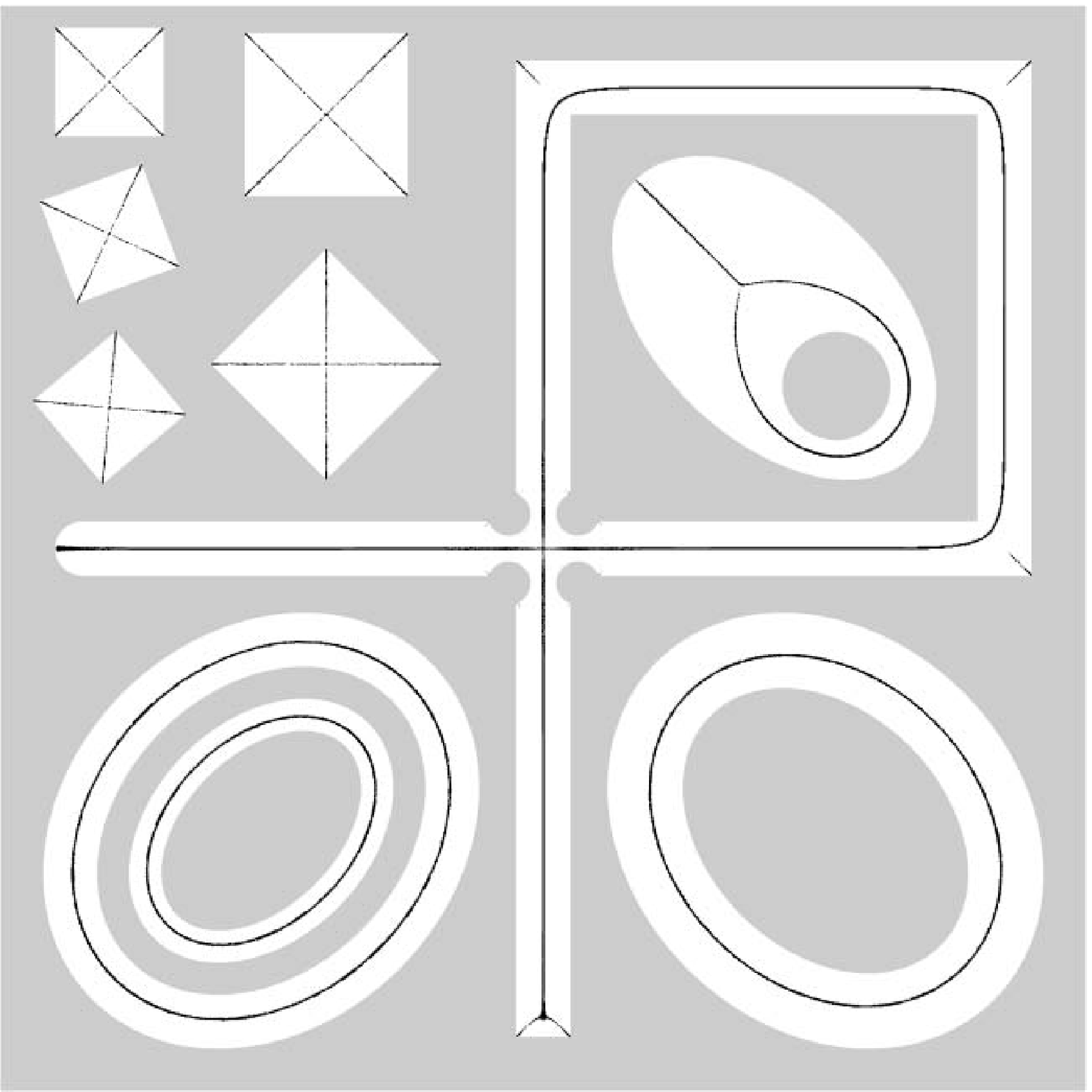}}
		\caption{Numerical example of a two-dimensional complex shape}
		\label{fig:2dResult2}
	\end{center}
\end{figure*}

An image with characteristic shapes is initially considered, as shown in Figure \ref{fig:2dResult2}.
Ring shapes with constant thicknesses located in the bottom portion of the images are considered. Therefore, the inverse thickness and thickness function values must be constant for each ring shape. 
It is confirmed that this requirement and the expected magnitude of the thickness function $h_f$ are satisfied.
In addition, the orientation vector and skeleton function also indicate appropriate features.

Cubic shapes located in the upper left corner are also considered here. Although each location and angle are different, the thickness function values are equivalent. Thus, it is confirmed that the dependency of these locations and angles are extremely low in the proposed method.
In addition, the orientation vectors and skeleton function indicate appropriate features. Note that the medial axes of the cubic shapes are its diagonal lines.

Next, the study focused on the cross shape located in the center of the image.
The intersection is made to provide a constant thickness in diagonal directions. 
It was confirmed that the thickness function value at the intersection was appropriate. That is, the thickness is equivalently evaluated as straight bars.

Finally, the full image is considered. The shape and topology of the image are extremely complex. 
However, all shape features are appropriately extracted simultaneously.
That is, the proposed method does not have any topological constraint.
%
%-----------------------
\subsection{General shapes}
The effectiveness of the proposed method for general shapes was examined.
For all examples, the image size is set to $1\times 1$ and the parameters of the PDE are set to $h_0=0.3$ and $a=0.2$. The reference domain is discretized using $\PP 2$ triangular finite elements.
\begin{figure*}[htb]
	\begin{center}
		\subfigure[Input image data]{
			\includegraphics[height=3cm]{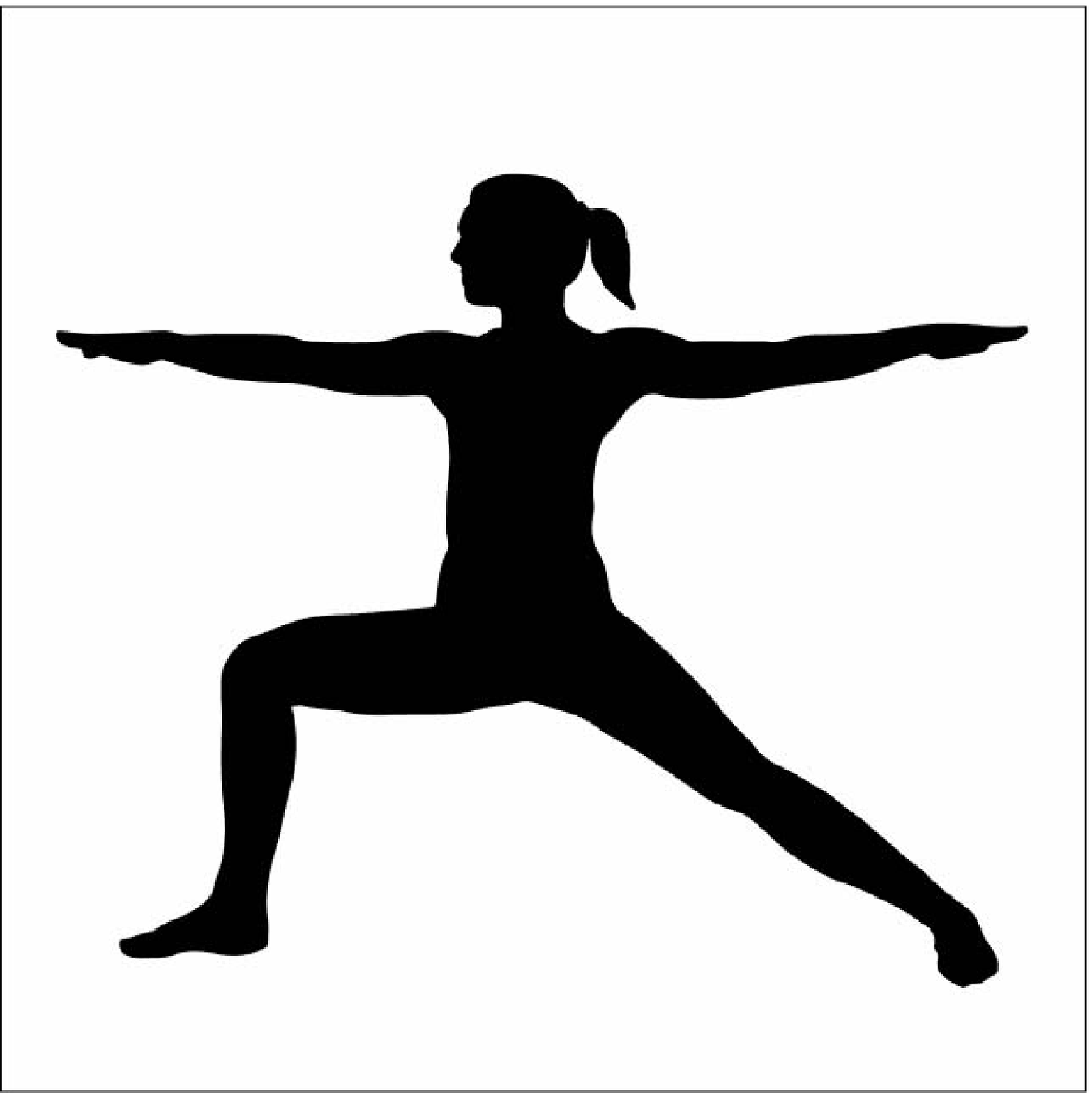}}\quad
		\subfigure[Inverse thickness function $f_h (s_1, s_2)$]{
			\includegraphics[height=3cm]{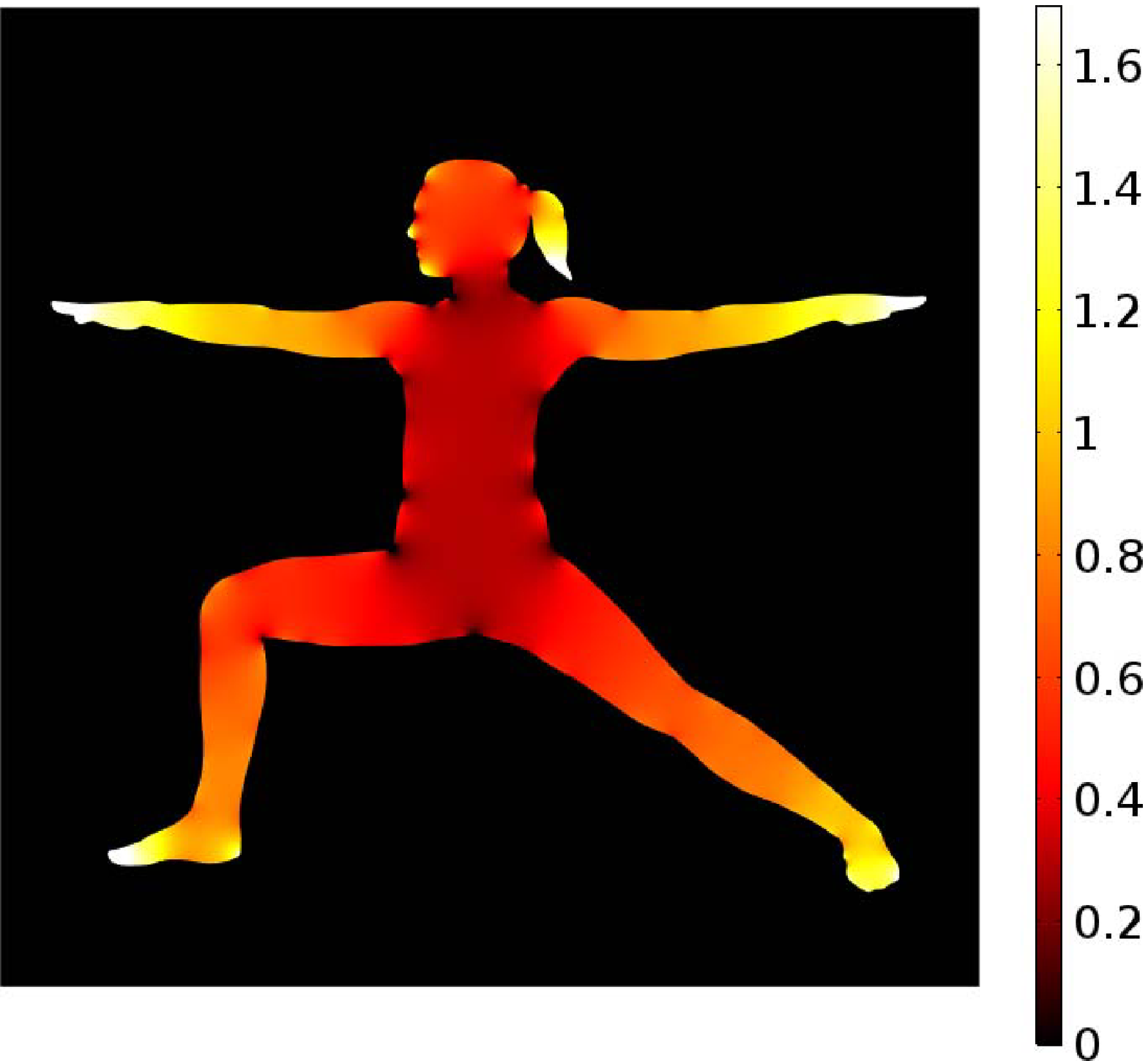}}\quad
		\subfigure[Thickness function $h_f (s_1, s_2)$]{
			\includegraphics[height=3cm]{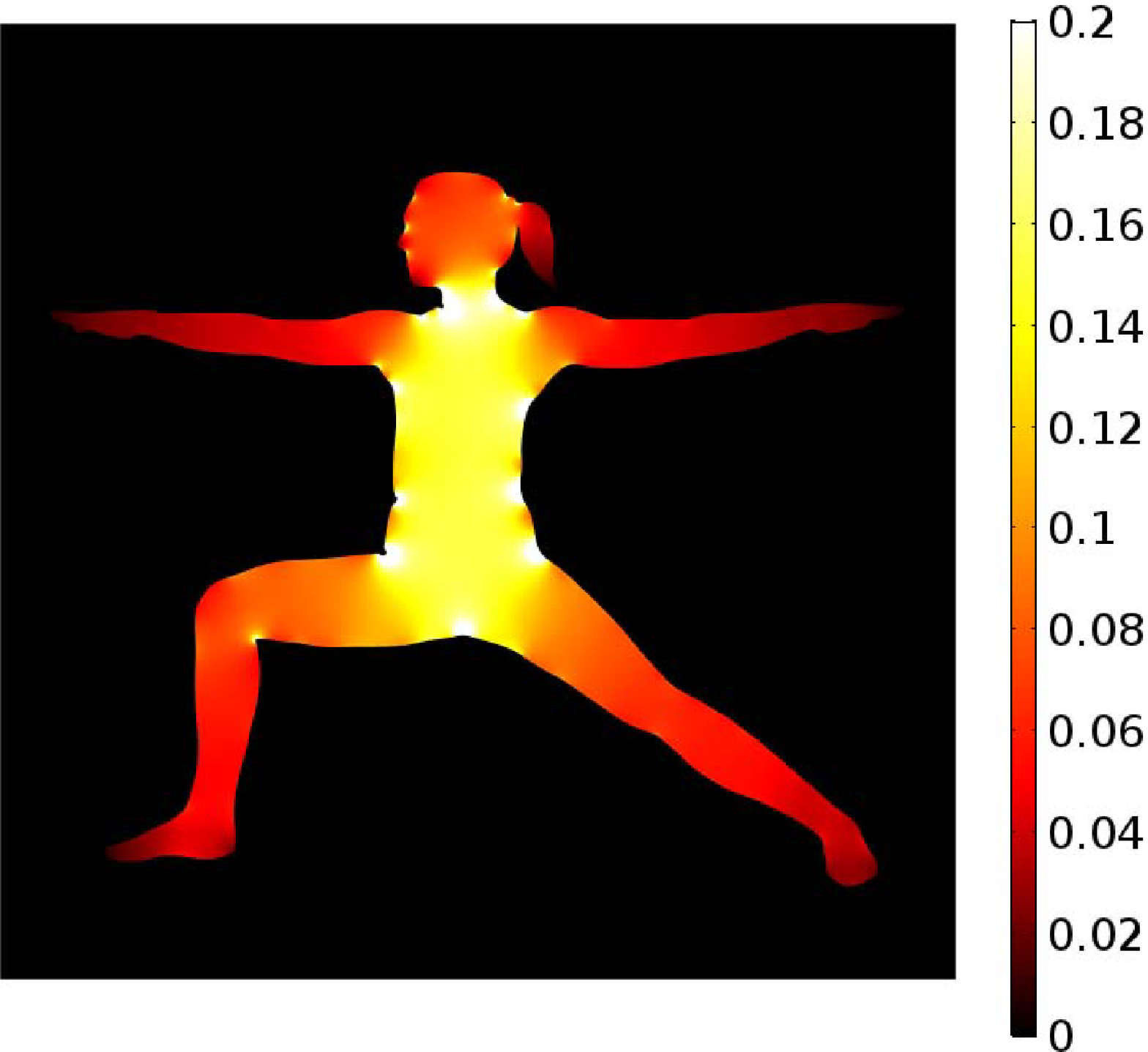}}\quad
		\subfigure[Orientation vectors]{
			\includegraphics[height=3cm]{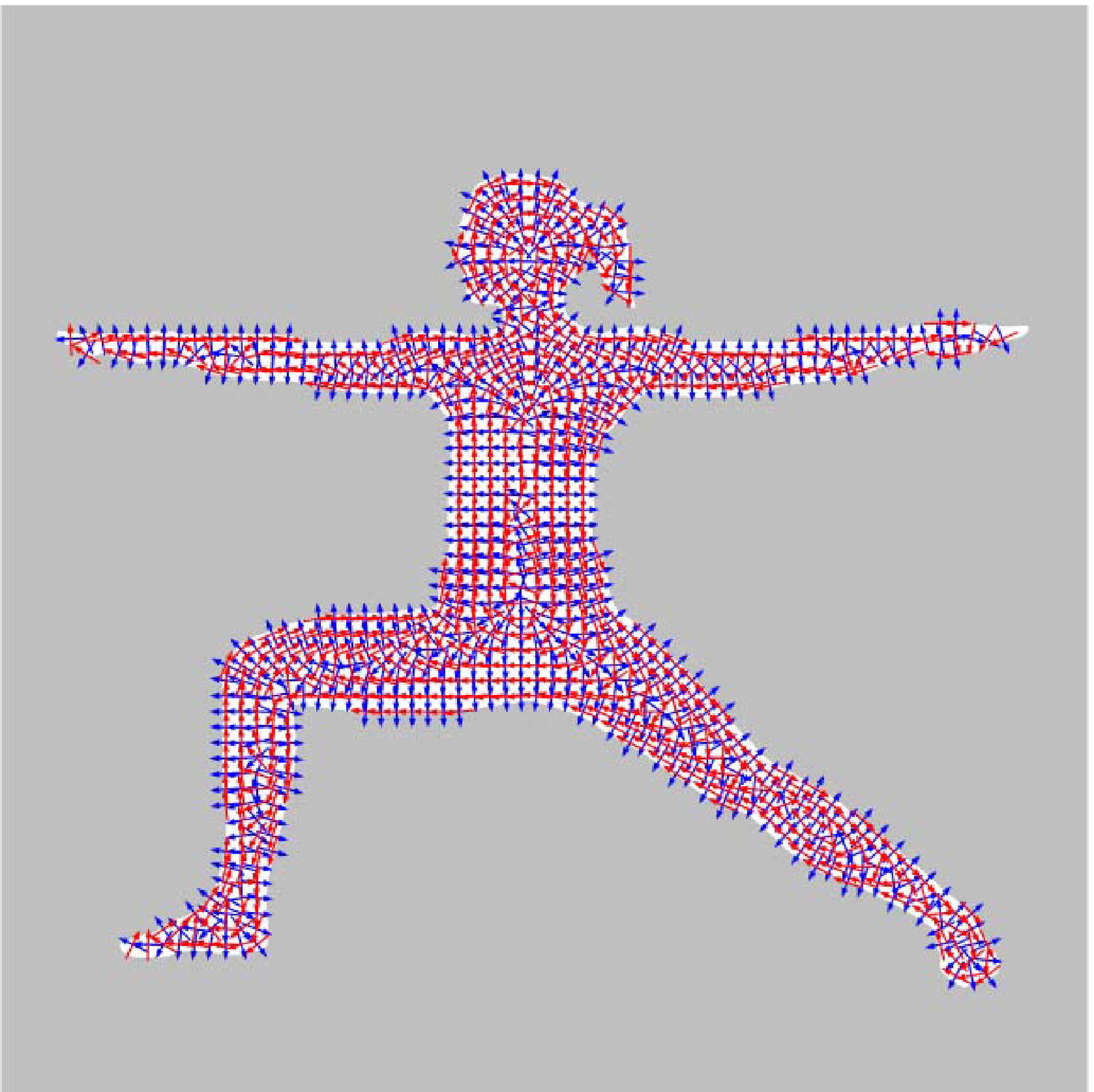}}\quad
		\subfigure[Skeleton function $f_s (s_1, s_2)$.]{
			\includegraphics[height=3cm]{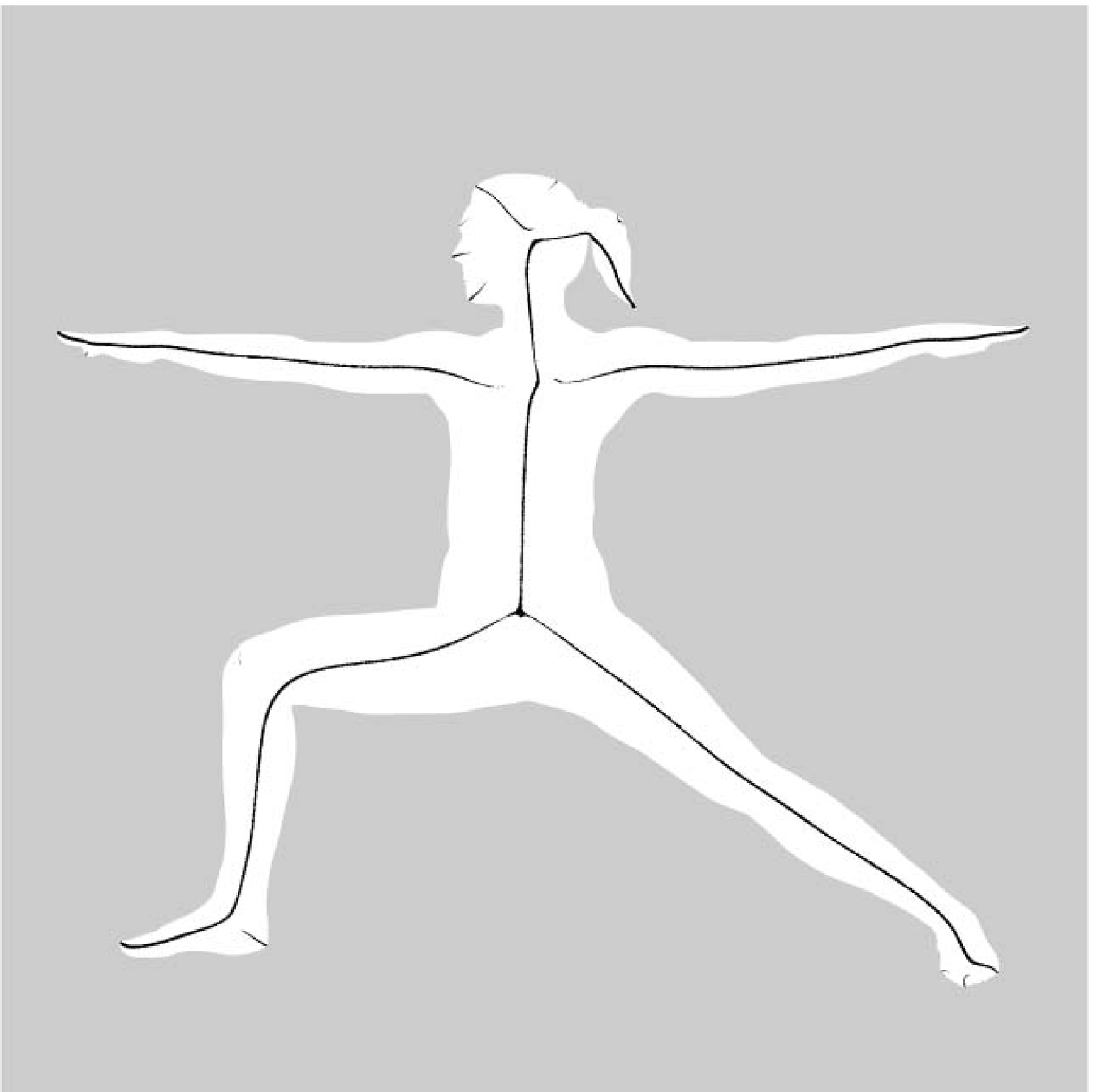}}
		\caption{Case 1 of a general shape in two dimensions}
		\label{fig:2dG1}
	\end{center}
\end{figure*}
%------------------------
\begin{figure*}[htb]
	\begin{center}
		\subfigure[Input image data]{
			\includegraphics[height=3cm]{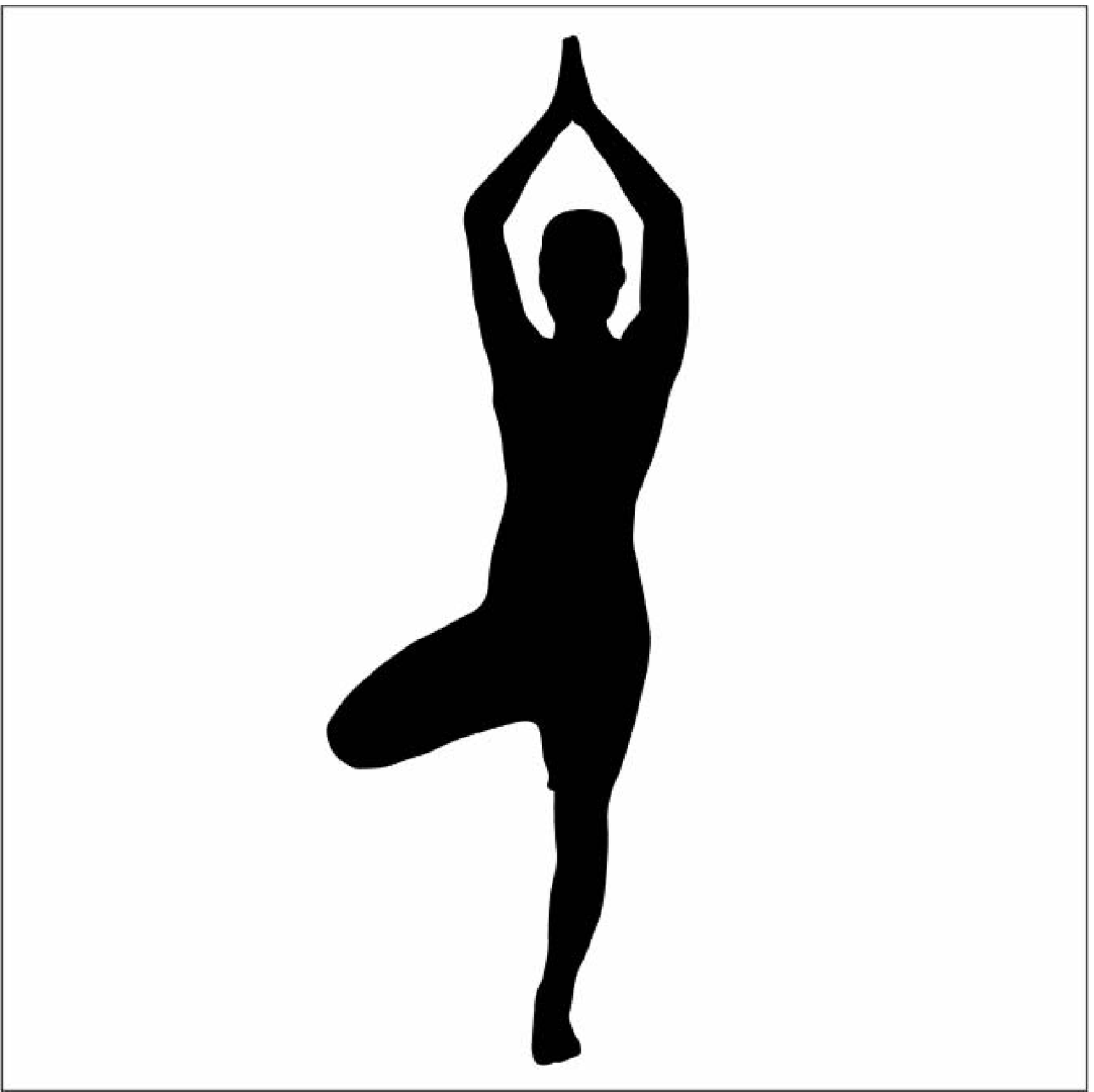}}\quad
		\subfigure[Inverse thickness function $f_h (s_1, s_2)$]{
			\includegraphics[height=3cm]{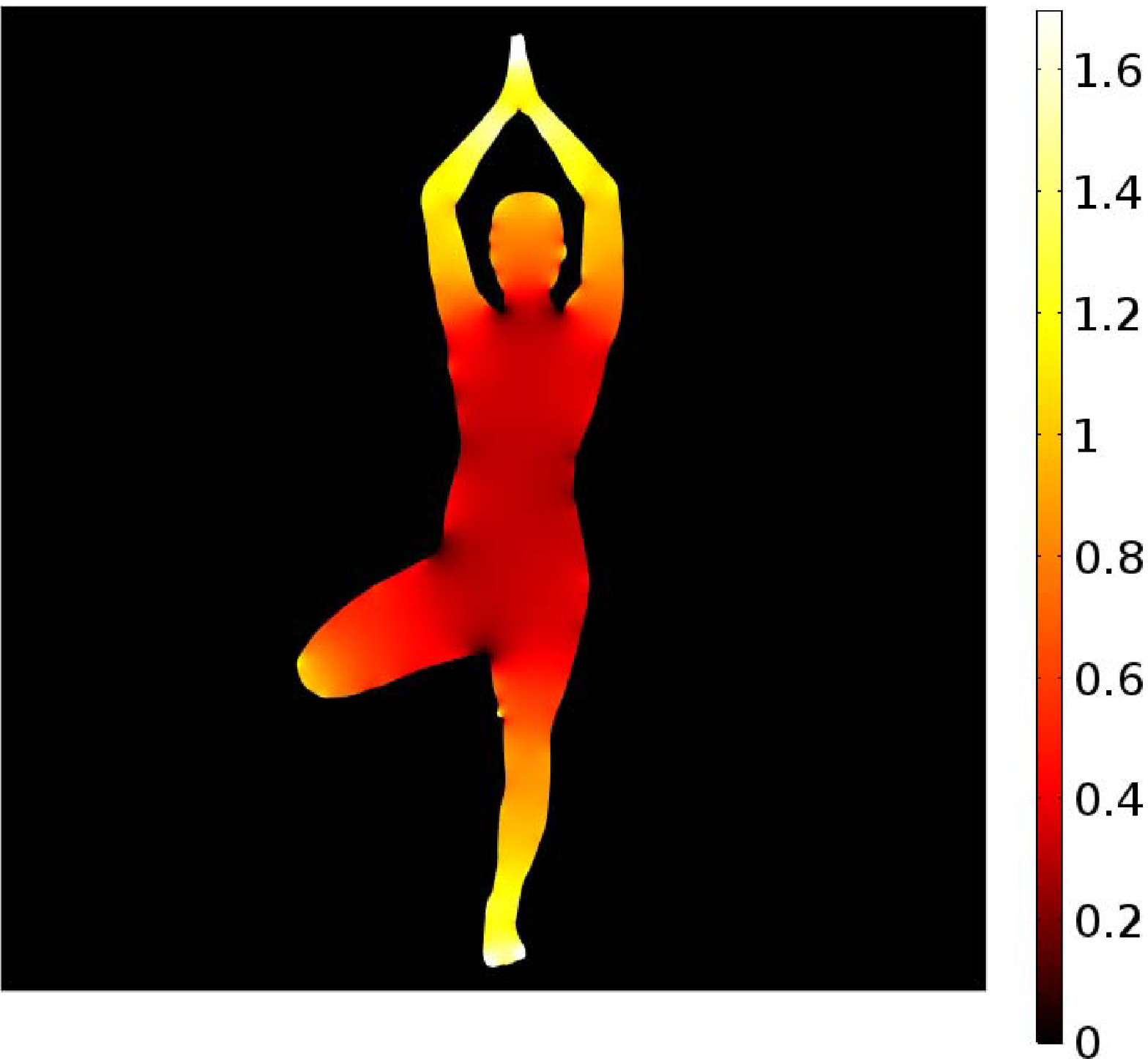}}\quad
		\subfigure[Thickness function $h_f (s_1, s_2)$]{
			\includegraphics[height=3cm]{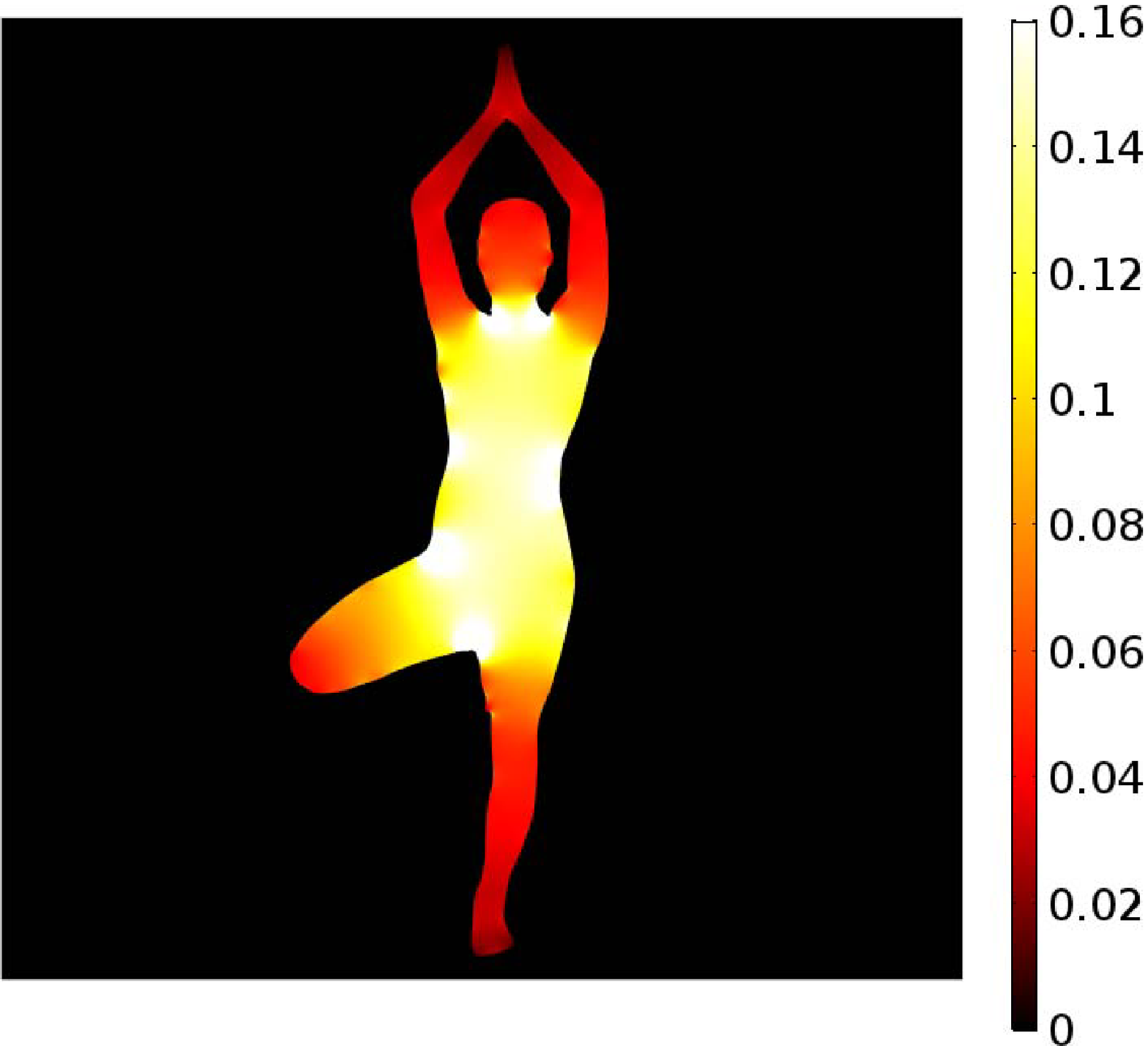}}\quad
		\subfigure[Orientation vectors]{
			\includegraphics[height=3cm]{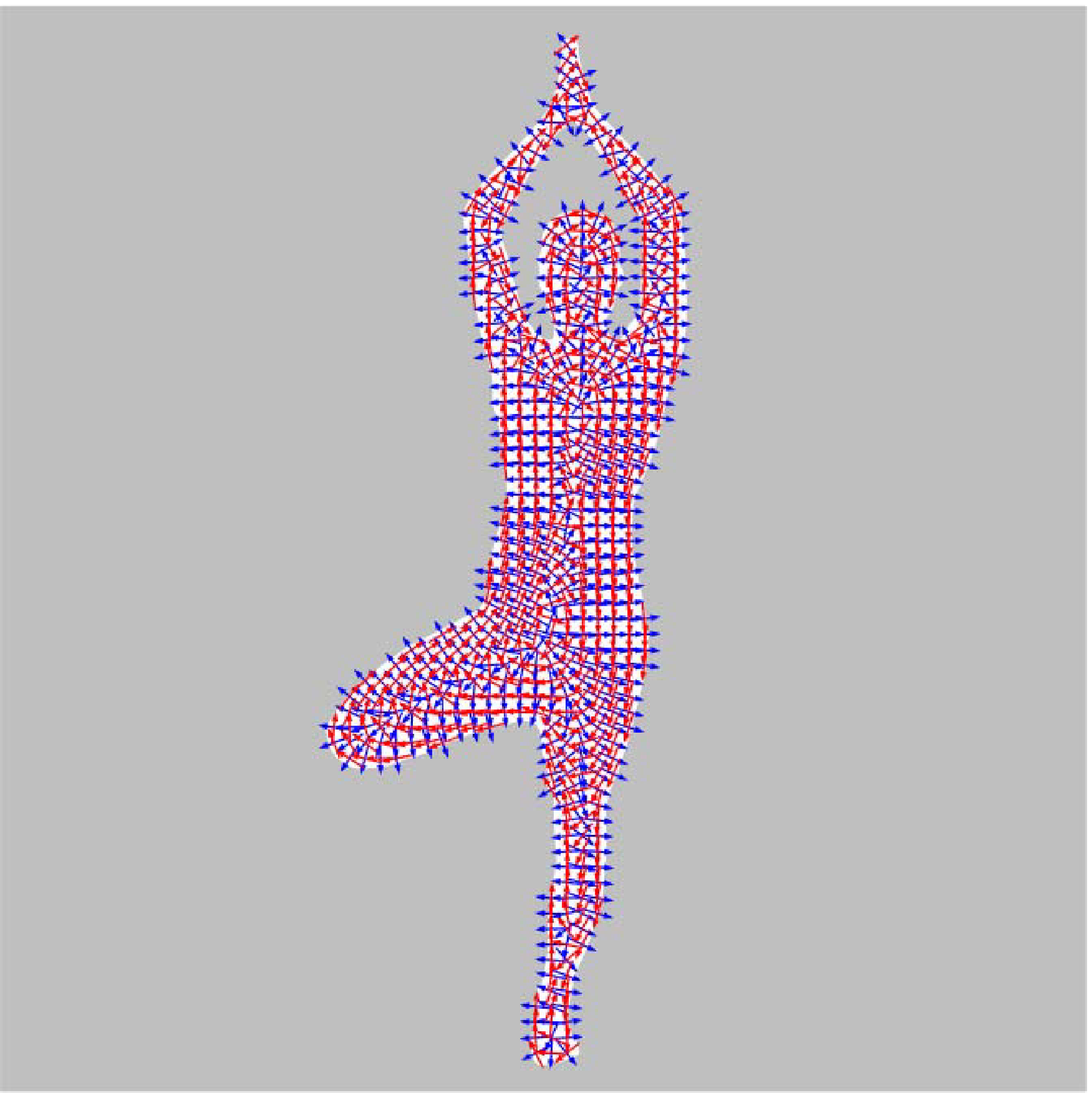}}\quad
		\subfigure[Skeleton function $f_s (s_1, s_2)$.]{
			\includegraphics[height=3cm]{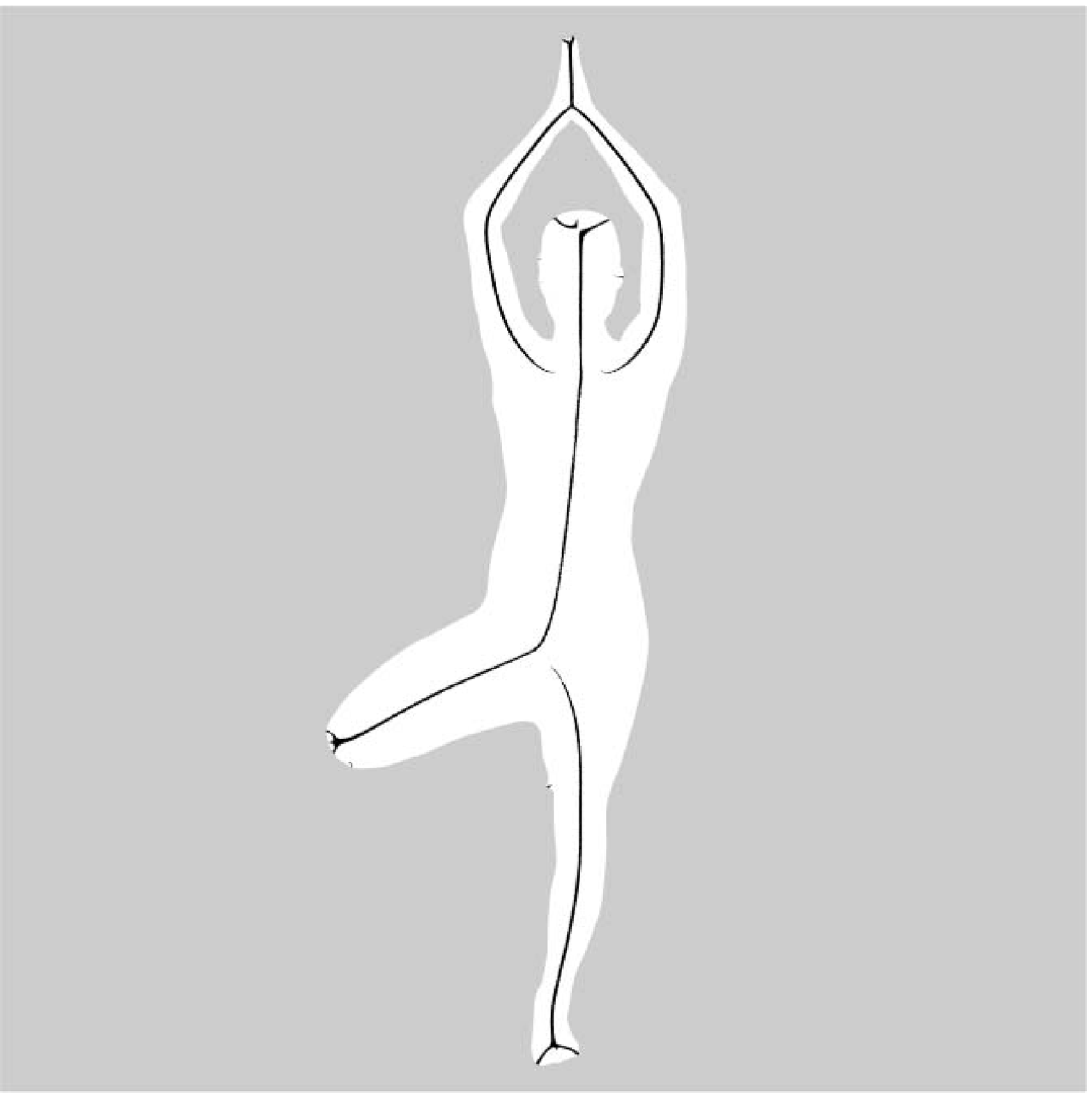}}
		\caption{Case 2 of a general shape in two dimensions}
		\label{fig:2dG2}
	\end{center}
\end{figure*}
%------------------------
\begin{figure*}[htb]
	\begin{center}
		\subfigure[Input image data]{
			\includegraphics[height=3cm]{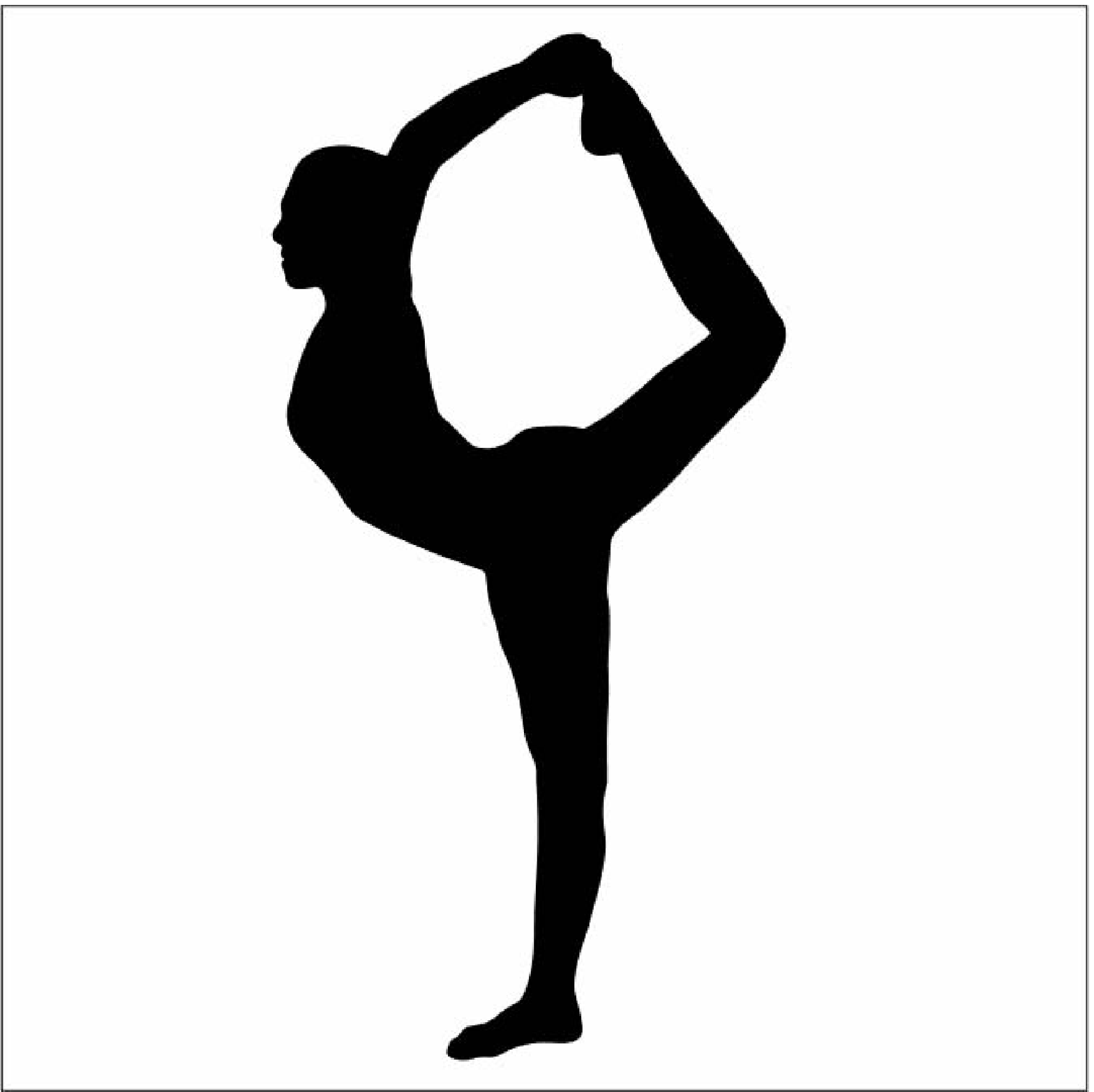}}\quad
		\subfigure[Inverse thickness function $f_h (s_1, s_2)$]{
			\includegraphics[height=3cm]{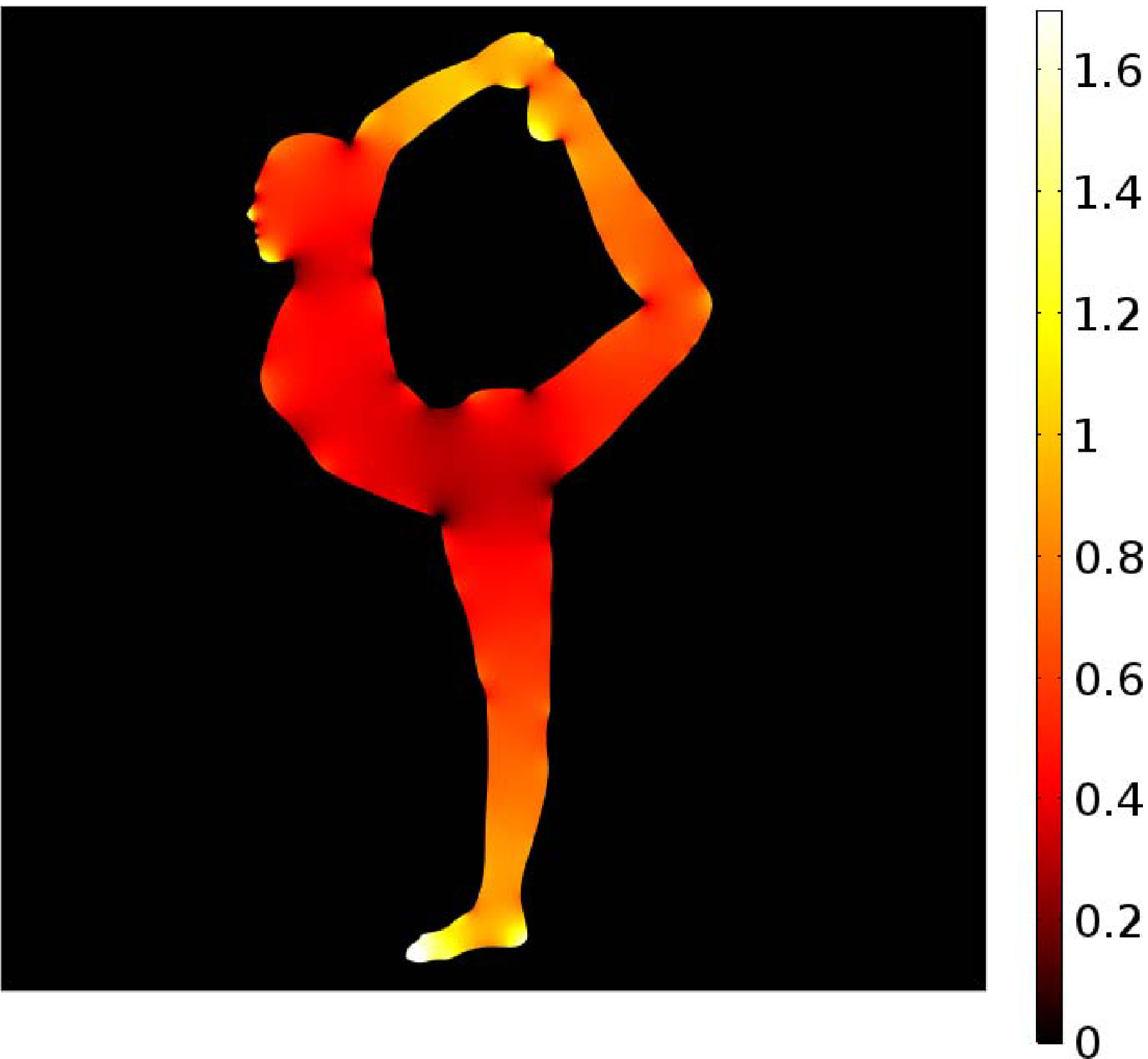}}\quad
		\subfigure[Thickness function $h_f (s_1, s_2)$]{
			\includegraphics[height=3cm]{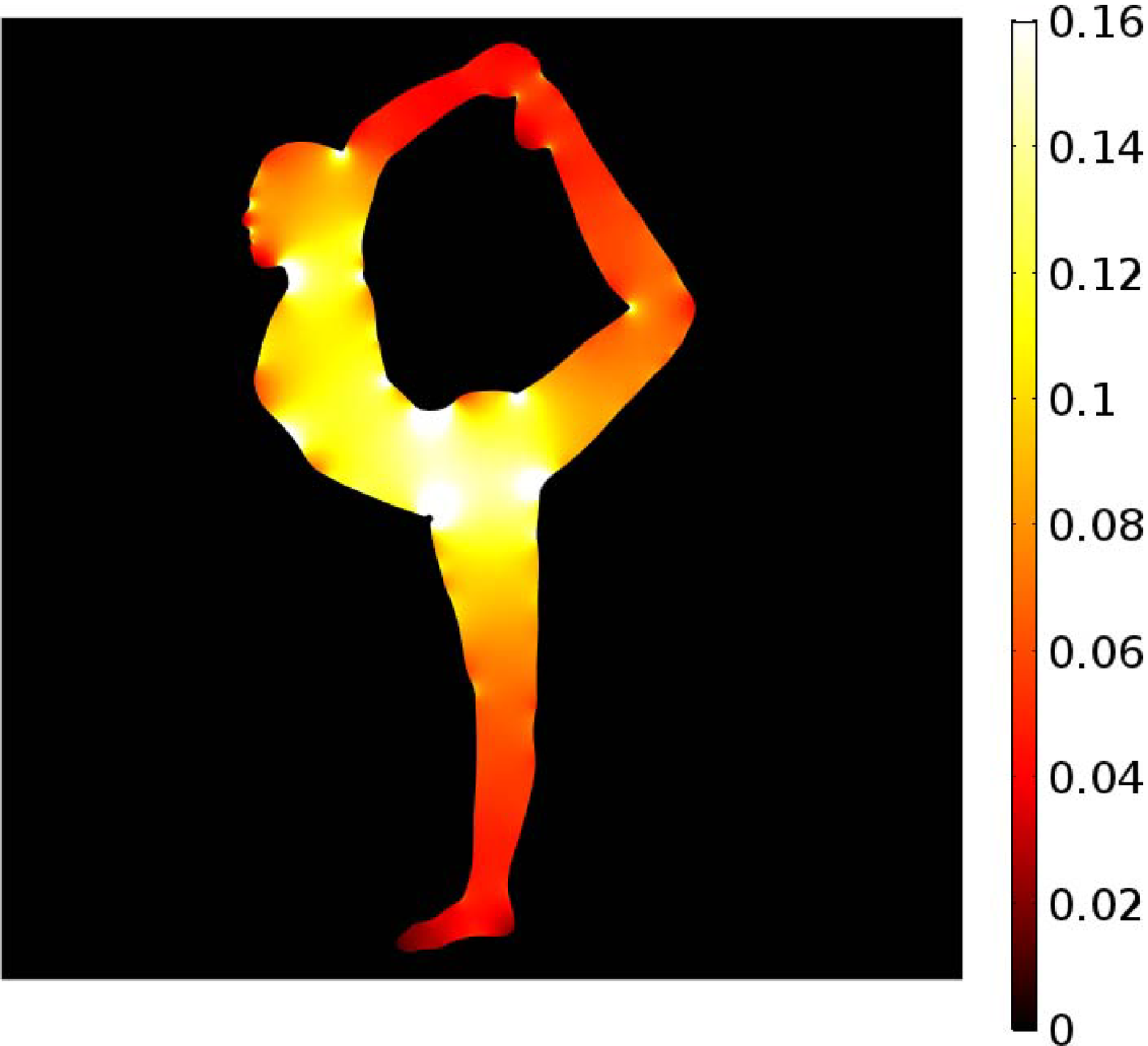}}\quad
		\subfigure[Orientation vectors]{
			\includegraphics[height=3cm]{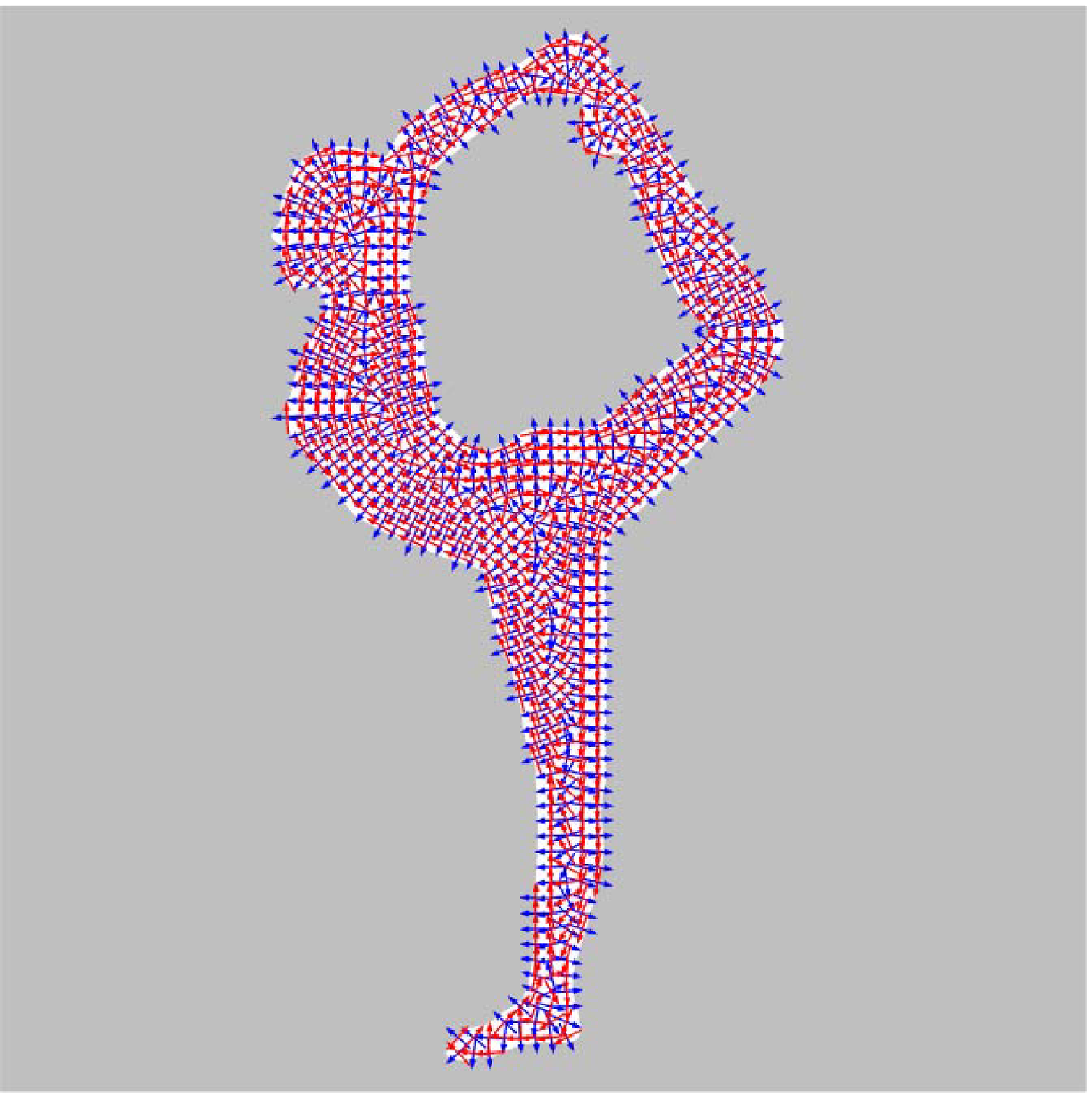}}\quad
		\subfigure[Skeleton function $f_s (s_1, s_2)$.]{
			\includegraphics[height=3cm]{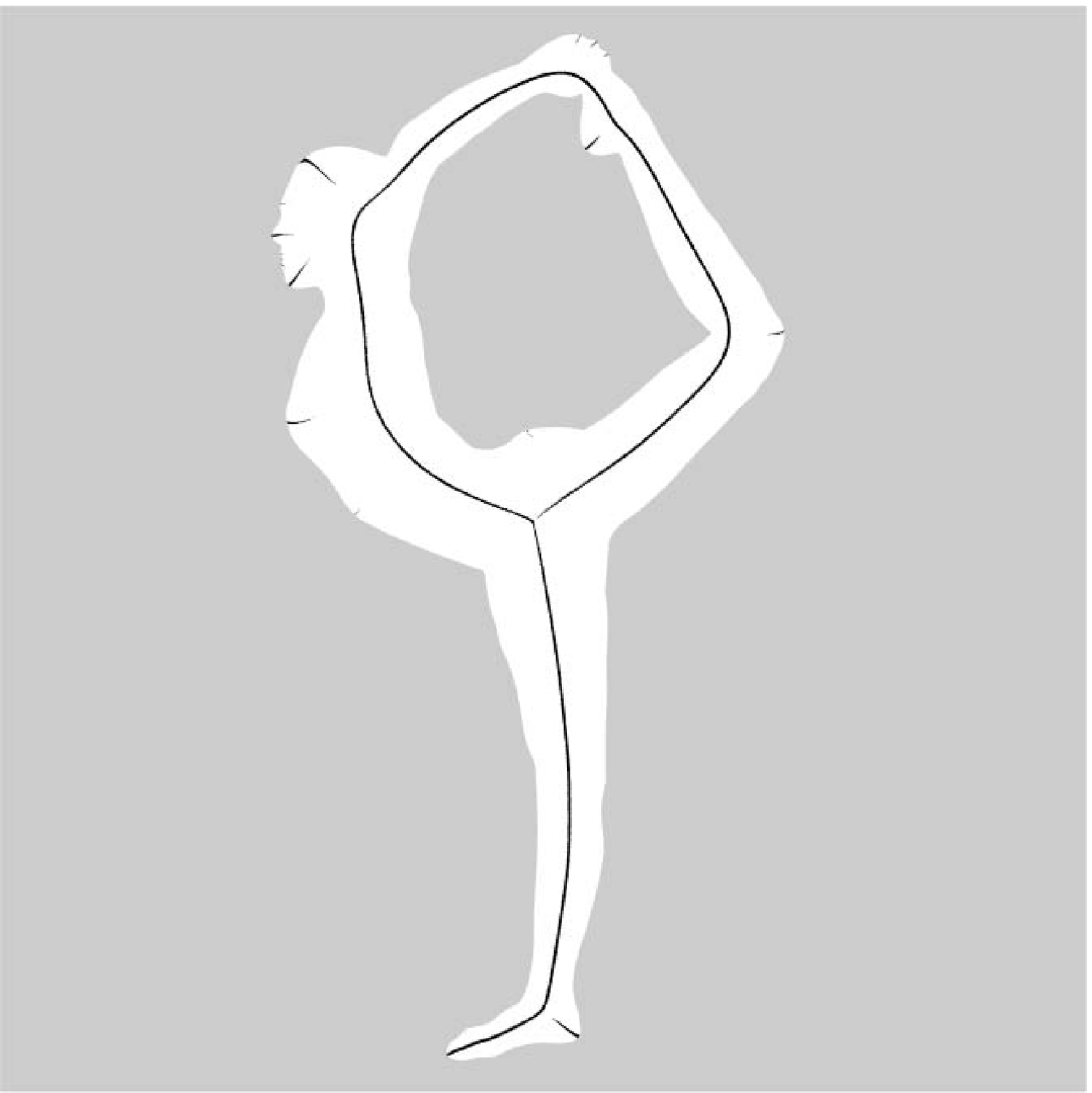}}
		\caption{Case 3 of a general shape in two dimensions}
		\label{fig:2dG3}
	\end{center}
\end{figure*}
Three general two dimensional shapes are considered here, as shown in Figures \ref{fig:2dG1}, \ref{fig:2dG2}, and \ref{fig:2dG3}.
As shown, the input images have complex geometric features.
The inverse thickness and thickness function values are globally appropriate, as shown in these figures.
From a local perspective, sharply dented shapes were estimated as thick shapes.
In addition, a relatively small fluctuated shape is neglected because the small feature is averaged by the diffusion effect in the PDE.
The orientation vectors and skeleton are also globally appropriate.
However, disconnected skeletons are obtained, because the proposed function is defined based on the concept of the medial axis, and connectivity is not considered. Therefore, different functions must be considered.

%-----------------------
\subsection{Tari's Dataset}
\begin{figure*}[htb]
	\begin{center}
		\subfigure[Inverse thickness function $f_h (s_1, s_2)$]{
			\includegraphics[height=4cm]{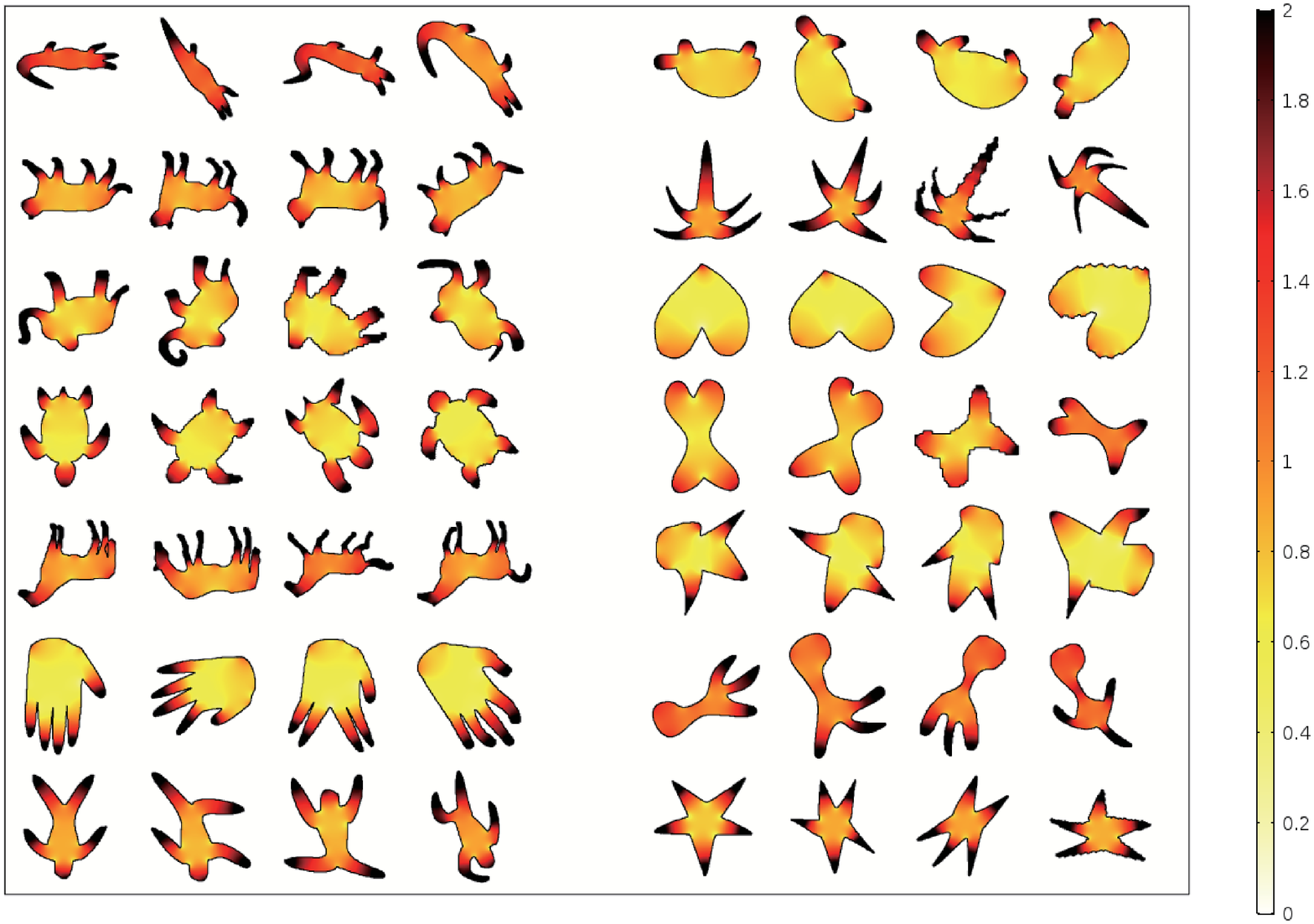}}\quad
		\subfigure[Thickness function $h_f (s_1, s_2)$]{
			\includegraphics[height=4cm]{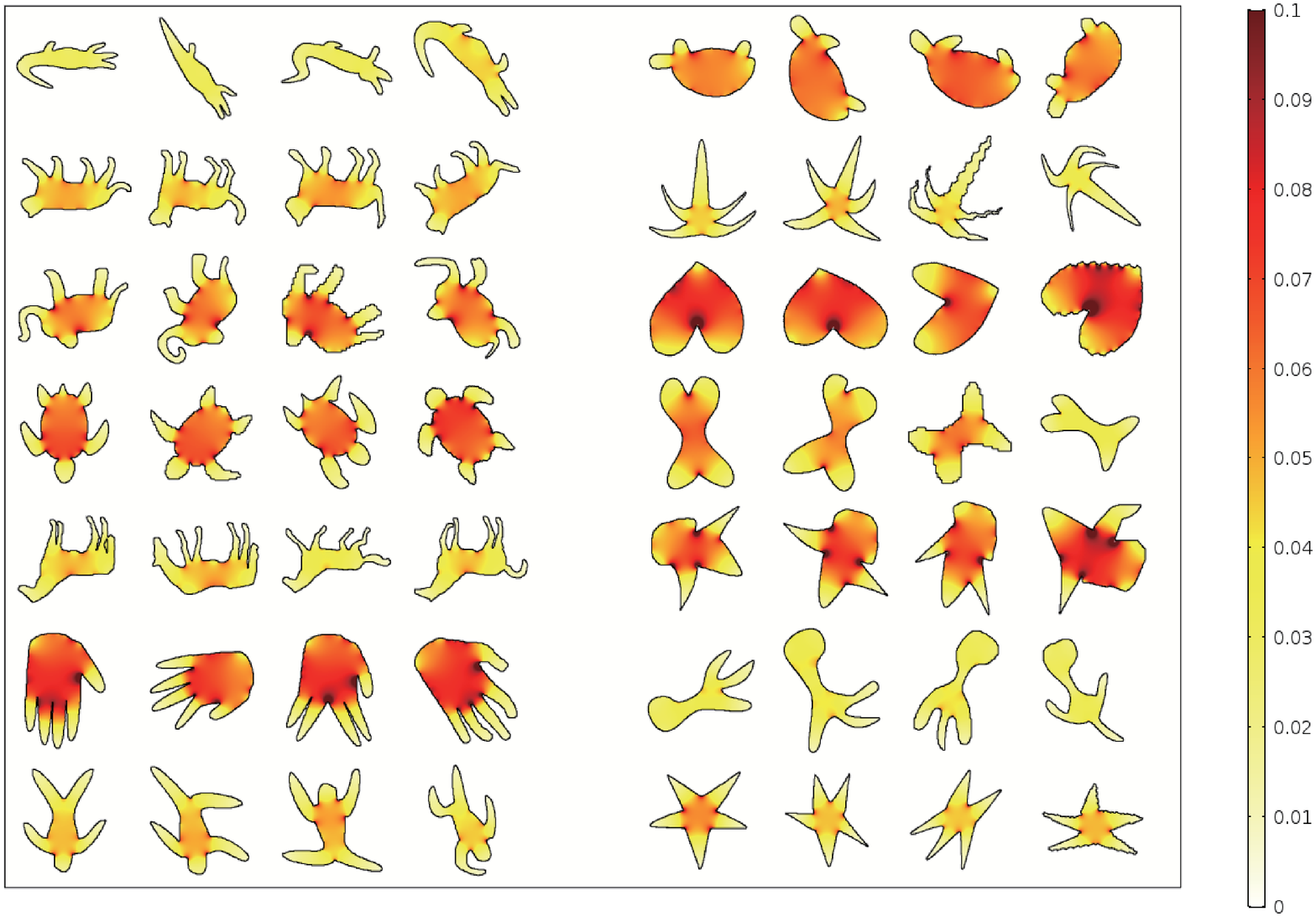}}\quad
		\subfigure[Orientation vectors on the shape boundaries]{
			\includegraphics[height=4cm]{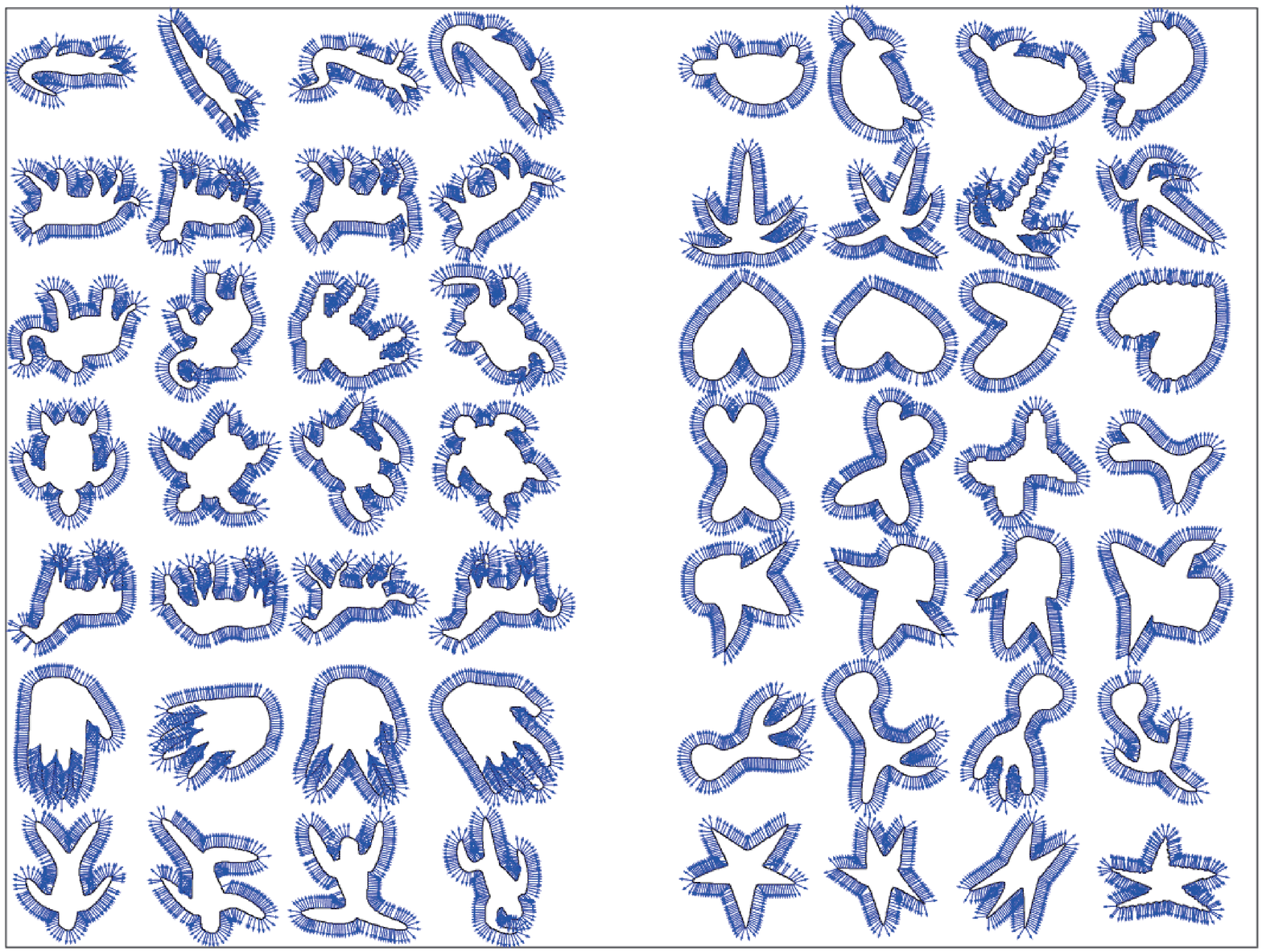}}\quad
		\subfigure[Skeleton function $f_s (s_1, s_2)$.]{
			\includegraphics[height=4cm]{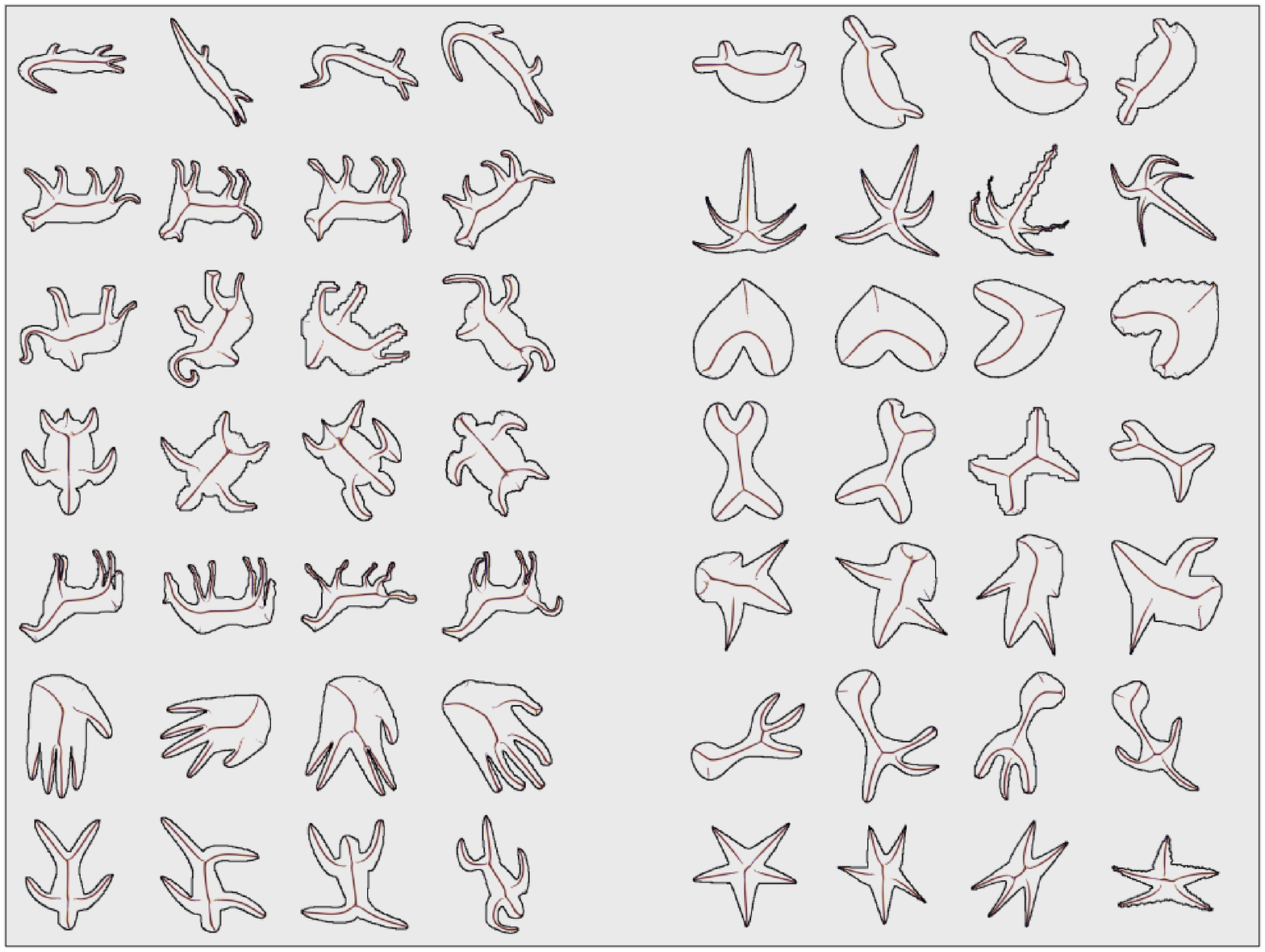}}
		\caption{Results with Taris' dataset using the proposed method}
		\label{fig:tari}
	\end{center}
\end{figure*}

%------------------------
All shapes in Tari's dataset \cite{asian2005axis} were compared with the related research. The dataset includes noisy images that are used as benchmark shapes in skeleton extraction \cite{gao20182d,shen2011skeleton,shen2013skeleton,aslan2008disconnected}.
The image size is set to $2\times 1.5$ in Figure \ref{fig:tari}. The parameters of the PDE are set to $h_0=0.05$ and $a=0.2$. The reference domain is discretized using $\PP 2$ triangular finite elements.
The obtained results are consistent with results from the literature \cite{gao20182d,shen2011skeleton,shen2013skeleton,aslan2008disconnected}. However, the obtained skeleton is disconnected. Therefore, the proposed skeleton extraction results are similar to Aslan's results \cite{aslan2008disconnected}.
%color end
%------------------------------------------------------------------
%--------------------------------------------------------------------------------------
% Section 
%--------------------------------------------------------------------------------------
%--------------------------------------------------------------------------------------

%------------------------------------------------------------------
%--------------------------------------------------------------------------------------
% Section 
%--------------------------------------------------------------------------------------
%--------------------------------------------------------------------------------------
\section{Conclusions and future work}
This paper presents a unified method for extracting geometric shape features using a steady state PDE.
The following results were obtained and the conclusions were drawn:
\begin{enumerate}
	\item A steady state PDE was formulated for extracting geometric features. 
	\item The functions of the orientation vector, inverse thickness, thickness, and skeleton were formulated as the solutions to the proposed PDE.
	\item The analytical solution was derived for the one-dimensional case. The derived solution demonstrates the validity of the proposed function.
	\item Several numerical examples were presented to confirm the usefulness of the proposed method for the various geometric shape features examined. In addition, it was confirmed that these geometric shape features were extracted without requiring any topological constraint.
\end{enumerate}

Functions for other geometric features, such as curved skeleton, will be mathematically considered in the future. In particular, heuristic formulations with respect to the thickness and the skeleton are required for a more precise extraction. The formulation will also be extended to gray images to expand the range of applications.

\section*{Acknowledgment}
	The author acknowledges comments from Professor Gr{\'e}goire Allaire ({\'E}cole Polytechnique) regarding validation of the proposed model.
	This work was supported in part by research grants from The Kyoto Technoscience Center and JSPS KAKENHI Grant Number 16K05039.

%\section*{References}

\bibliography{refs3}

\end{document}